\documentclass{IEEEtaes}

\usepackage{color,array,amsthm}
\usepackage{graphicx}
\usepackage{amssymb}
\usepackage{amsmath}
\usepackage{amsthm}
\usepackage{cases}
\usepackage{multirow}
\usepackage{longtable,tabularx}
\usepackage{subfig}
\newcommand{\sgn}{\text{sgn}}
\jvol{XX}
\jnum{XX}
\jmonth{XXXXX}
\paper{1234567}
\pubyear{2024}
\doiinfo{TAES.2024.Doi Number}
\newtheorem{theorem}{Theorem}
\newtheorem*{remark}{Remark}

\begin{document}
\title{Quadrotor Guidance for Window Traversal: A Bearings-Only
	Approach} 
\twocolumn[
\noindent\fbox{%
	\parbox{\dimexpr\columnwidth-2\fboxsep-2\fboxrule\relax}{%
		\centering \footnotesize
		This work has been submitted to the IEEE for possible publication. Copyright may be transferred without notice, after which this version may no longer be accessible.
	}%
}
\vspace{1em} 
]

\author{MIDHUN E K}

\author{ASHWINI RATNOO}

\affil{Indian Institute of Science, Bengaluru, 560012, India}

\authoraddress{Authors' addresses: Midhun E K and Ashwini ratnoo are with Indian Institute of Science, Bengaluru, 560012, India, E-mail:  
(\href{mailto:midhunlbs@gmail.com}{midhunlbs@gmail.com}; \href{mailto:ratnoo@iisc.ac.in}{ratnoo@iisc.ac.in}). {\itshape(Corresponding author: Midhun E K.)}}

\markboth{MIDHUN AND RATNOO}{QUADROTOR GUIDANCE FOR WINDOW TRAVERSAL}
\maketitle

\begin{abstract}This paper focuses on developing a bearings-only measurement-based three-dimensional window traversal guidance method for quadrotor Uninhabitated Aerial Vehicles (UAVs). The desired flight path and heading angles of the quadrotor are proposed as functions of the bearing angle information of the four vertices of the window. These angular guidance inputs employ a bearing angle bisector term and an elliptic shaping angle term, which directs the quadrotor towards the centroid of the window. Detailed stability analysis of the resulting kinematics demonstrates that all quadrotor trajectories lead to the centroid of the window along a direction which is normal to the window plane. A qualitative comparison with existing traversal methodologies showcases the superiority of the proposed guidance approach with regard to the nature of information, computations for generating the guidance commands, and flexibility of replanning the traversal path. Realistic simulations considering six degree-of-freedom quadrotor model and Monte Carlo studies validate the effectiveness, accuracy, and robustness of the proposed guidance solution. 
	Representative flight validation trials are carried out using an indoor motion capture system.
\end{abstract}

\section{INTRODUCTION}
W{\scshape ith} autonomous capabilities, small size, maneuver agility, and low cost, quadrotors are increasingly used for a wide range of complex real-world missions such as surveillance, inspection, autonomous drone racing, and search-and-rescue operations in densely populated areas. To successfully execute these missions, quadrotors need to autonomously traverse through gaps while flying in cluttered environments characterized by obstacles, physical boundaries, and confined spaces. In doing so, they commonly encounter traversal through rectangular windows. Feasible guidance methods are needed for successfully and safely executing such window traversal maneuvers. 
While developing guidance strategies in real-world scenarios, it is essential to consider the type of information provided by onboard sensors.
 Due to the inherent payload limitations of flying vehicles, a popular choice is to use passive, lightweight, and inexpensive vision-based sensors \cite{visonsurvey}. Relative bearing information of the image features can be readily obtained using intrinsic parameters of a vision-based sensor \cite{ma2005invitation}, which is valuable for developing traversal guidance solutions.
 
Guidance laws utilizing bearing information are widely reported in the context of interceptor-target engagements \cite{pz3,doi:10.2514/1.G002459,bpn,yashbearing,LEE2010136}.
Typical guidance objective in those works is to intercept a stationary or a moving target. Guidance laws based only on bearing information have also gained some attention in the context of small UAV applications. The authors in \cite{doi:10.2514/6.2015-0074} and \cite{PN} introduce a guidance strategy for UAV collision avoidance based on proportional navigation guidance law.
Therein, bearing rates obtained from a monocular vision sensor are used to identify the collision threats. Further, guided maneuvers vary the magnitude of bearing rate to avoid collisions. 
Trinh et al. \cite{3beacontriangular} propose a guidance method to direct a quadrotor to any desired location within a triangular area utilizing bearing information from three stationary noncollinear beacons. That method is further extended in \cite{3beaconplane} to enable the vehicle to access any location across the entire plane. In \cite{shalini}, a gain adaptive proportional navigation approach is developed for wall-following application, utilizing relative bearing information from beacons placed along the wall. References \cite{formation,formation2} utilize inter-agent bearing information for  formation control of quadrotors, enabling them to achieve a desired geometric configuration. Another work considers a pursuer-quadrotor using three-dimensional bearing information-based guidance commands for following a target-quadrotor by maintaining a specified relative distance \cite{helical}. Bearing information-based quadrotor guidance strategy is also employed in the context of lane changing within an air corridor system \cite{lanechanging}. In that approach, the quadrotor utilizes bearing measurements of neighboring quadrotors in the destination lane to execute a safe lane change maneuver. 

Several methods have discussed guidance of a quadrotor flying through openings which are characterized by stationary physical boundaries.
 Mellinger et al. \cite{b3} developed concatenated trajectories and associated controllers that maneuver a quadrotor to a goal location while passing through a window.
 The sequence of trajectories include hovering to a desired starting position, a three-dimensional trajectory through the window, and finally a hover approach to the desired goal position. Mellinger and Kumar \cite{b4} propose a traversal algorithm generating optimal trajectory passing through a series of waypoints, while guaranteeing safe traversal through a predetermined gap. The quadrotor trajectory therein is represented by piecewise time-parameterized polynomials, the coefficients of which are determined by optimizing integral of the squared norm of the snap (second derivative of acceleration). A four-parameter logistic curve-based safe trajectory generation method is proposed in \cite{upadhyay2022generation} for three-dimensional traversal through a window. Another work \cite{MPCC} uses a sigmoid function-based geometric interpolation to facilitate the quadrotor traversal through a series of windows which are placed along two parallel lanes. It is noteworthy that Refs. \cite{b3,b4,upadhyay2022generation,MPCC} rely on prior known information about the dimensions of the gap and the final goal point in the world frame, which may not be available in all realistic scenarios. Reference \cite{mekjrnl} addressed a two-dimensional gap traversal problem, and the quadrotor's heading guidance input is adjusted based on the bearing orientations of the two gap extremities. 
However, realistic window traversal scenarios are typically three-dimensional and that work is not applicable in such problems.

Based specifically on vision-based guidance, several studies discuss quadrotor gap traversal problems \cite{estimation,activevision,Guo,QuadrotorGuidanceimage,imagematching,CNN2,CNN}. Loianno et al. \cite{estimation} address quadrotor traversal through a window using information obtained from an onboard vision sensor and an inertial measurement unit and the trajectory is generated in a fashion similar to the minimum snap approach in \cite{b4}. Therein, the position and orientation of the window with respect to the quadrotor's starting location are assumed to be known apriori. Further, the methods in \cite{b3,b4,upadhyay2022generation,estimation} generate complete traversal trajectory at once with inflexibility of using revised or updated information of the gap. 
With such an approach, errors in gap information can significantly affect the safety of the proposed trajectory. Falanga et al. \cite{activevision} consider quadrotor position profile as a quadratic function of time for it to pass through the geometric center of a gap, which is computed using feature points and depth information from the images.

References \cite{Guo,QuadrotorGuidanceimage} rely on prior window-image information-based trajectory generation, necessitating access to the original complete image of the window in advance. A polynomial trajectory generation method based on invariant features defined in image space is developed in \cite{Guo} by minimizing the snap in translation between the camera image and the reference image. Upon reaching the location where the original image is taken, a gap traversal trajectory is generated similar to the one in \cite{activevision}. Mueller et al. \cite{QuadrotorGuidanceimage} proposed a model predictive control framework for quadrotor guidance for window traversal. That work utilizes an image matching algorithm \cite{imagematching} to identify the target window within the current image. It is noteworthy that, the methods \cite{Guo,QuadrotorGuidanceimage} require depth information of the four window vertices by decomposition of homography matrix. 

In the context of qadrotor racing missions, Refs. \cite{CNN2,CNN} present learning-based solutions for traversal guidance. In \cite{CNN2}, a convolution network is used to estimate the window pose, and the quadrotor trajectory is generated by interpolating two waypoints—one prior to the window and the other beyond it. Another study \cite{CNN} utilizes convolution neural network to estimate the window center and a guidance law that uses the line-of-sight bearing angle of the window center for traversal.
Learning-based methods often require a large amount of data for training. The performance of the traversal model is heavily reliant on the quantity and quality of the training data, which may not be readily available in many unforeseen scenarios. Further, learning methods can be computationally expensive during training and deployment, and they may lack safety and performance guarantees.

This work proposes a new quadrotor guidance method for safely traversing through a rectangular window. The proposed approach relies solely on the three-dimensional bearing angles of the window vertices. 
Using the bisector of the angles, a desired flight path angle and a desired heading angle are introduced, which command the velocity components for the quadrotor. 
The proposed guidance commands are simple, expressed in closed form, and offer analytical trajectory convergence guarantees. 
Very preliminary findings of this research
were provided in \cite{mekcodit} focusing on sample simulation
results. However, that work did not consider a realistic quadrotor
model, uncertainties in bearing information, dynamic constraints of the
quadrotor attitude angles, stability analysis, qualitative comparison with existing methods, and experimental validation. This work addresses all those important gaps and establishes the merit of the proposed guidance method through a comprehensive research effort.

The rest of the paper is organized as follows: Section \ref{sec2} formulates the problem, and the proposed guidance solution is detailed in Section \ref{PM}. Stability characteristics are analyzed in Section \ref{stability}. Quadrotor dynamic modeling and control is discussed in Section \ref{dynamics}. Simulation results are presented in Section \ref{Numerical} and flight experiment results are discussed in Section \ref{Exprmnt}. Section \ref{Conclusion} provides conclusive remarks and future research directions.
\section{PROBLEM DESCRIPTION}\label{sec2}
\begin{figure}[htbp]
	\centerline{\includegraphics[width=83mm]{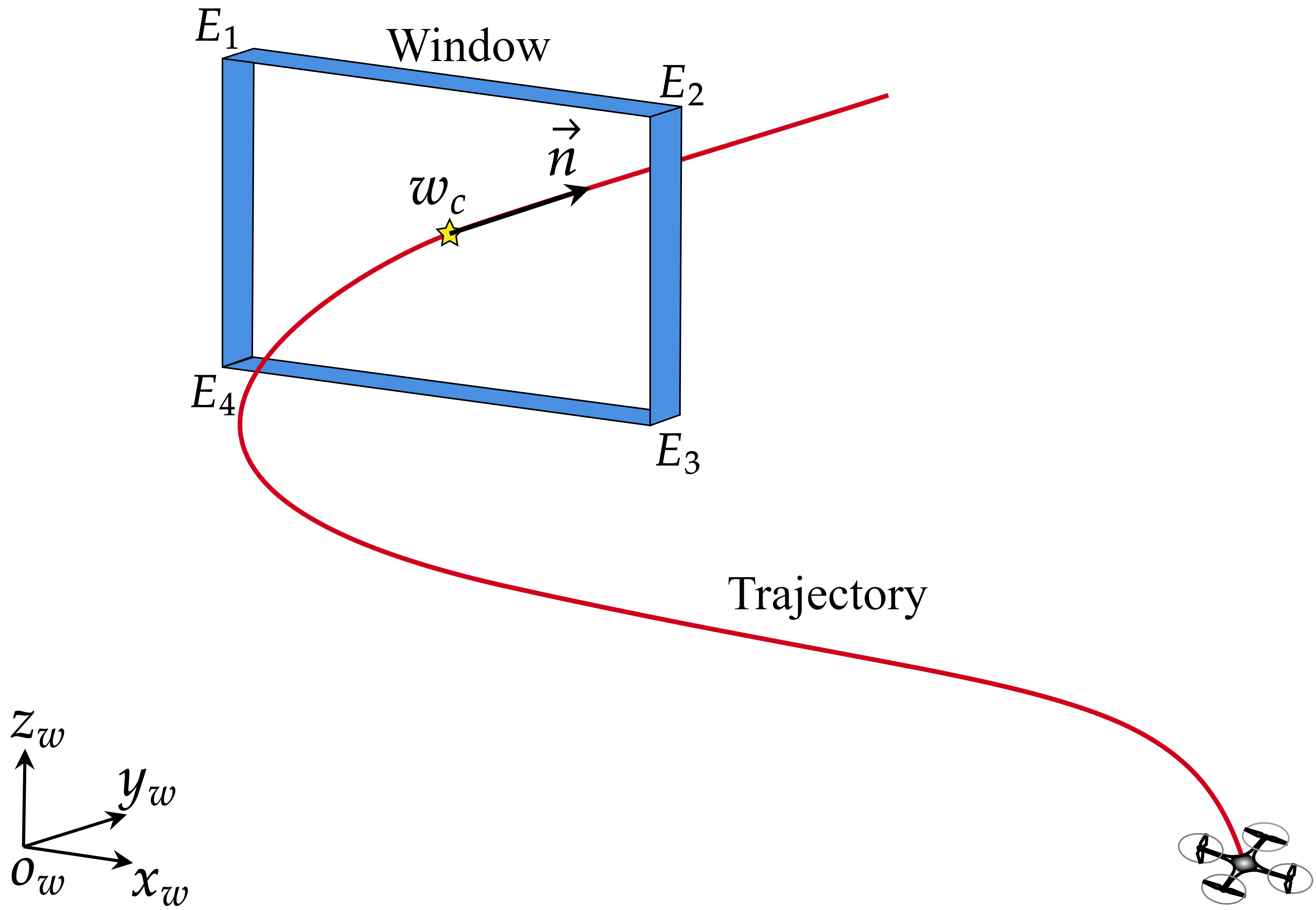}}
	\caption{Quadrotor traversal through a window.}
	\label{fig}
\end{figure}
Consider a three-dimensional scenario shown in Fig. \ref{fig}, where a point object quadrotor approaches a rectangular window.
Consider the world frame $\mathcal{W}=[x_w,y_w,z_w]$ with its origin at $o_w$ and plane $x_w-z_w$ parallel to the window plane. The four vertices of the window are denoted as $E_1$, $E_2$, $E_3$, and $E_4$. The position vectors corresponding to these four vertices in the world frame are represented as $[x_i\quad y_i\quad z_i]^T\in \mathbb{R}^3$, where $i=1,2,3$, and $4$. 
Let, $[\dot{x}\quad \dot{y}\quad \dot{z}]^T\in \mathbb{R}^3$ and $[x\quad y\quad z]^T\in \mathbb{R}^3$ represent the velocity and position vectors of the quadrotor in $\mathcal{W}$, respectively. 
And $V\in\mathbb{R}$, $\gamma\in\left[-\frac{\pi}{2},\frac{\pi}{2}\right]$, and $\chi\in[-\pi,\pi)$ represent speed, flight path angle, and heading angle of the quadrotor, respectively. Kinematic equations of quadrotor motion in three-dimensional space can be expressed as
\begin{align}\label{pointmass}
	\dot{x}(t) &=V \cos{\gamma(t)}\cos{\chi(t)}\\
	\dot{y}(t) &=V \cos{\gamma(t)} \sin\chi(t)\\
	\dot{z}(t) &=V \sin{\gamma(t)}\label{pointmassend}
\end{align} 
This work assumes that the quadrotor has an onboard vision sensor capable of measuring the three-dimensional bearing angles of four window vertices.
Using the information available, the objective is to design guidance solutions, $\gamma$ and $\chi$, that enable the quadrotor to fly through the centroid of the window $w_c$ as shown in Fig. \ref{fig}, along a direction $\vec{n}$, which is the normal vector into the window plane that passes through the $w_c$.
\section{PROPOSED TRAVERSAL GUIDANCE APPROACH}\label{PM}
\begin{figure}[htbp]
	\centerline{\includegraphics[width=73mm]{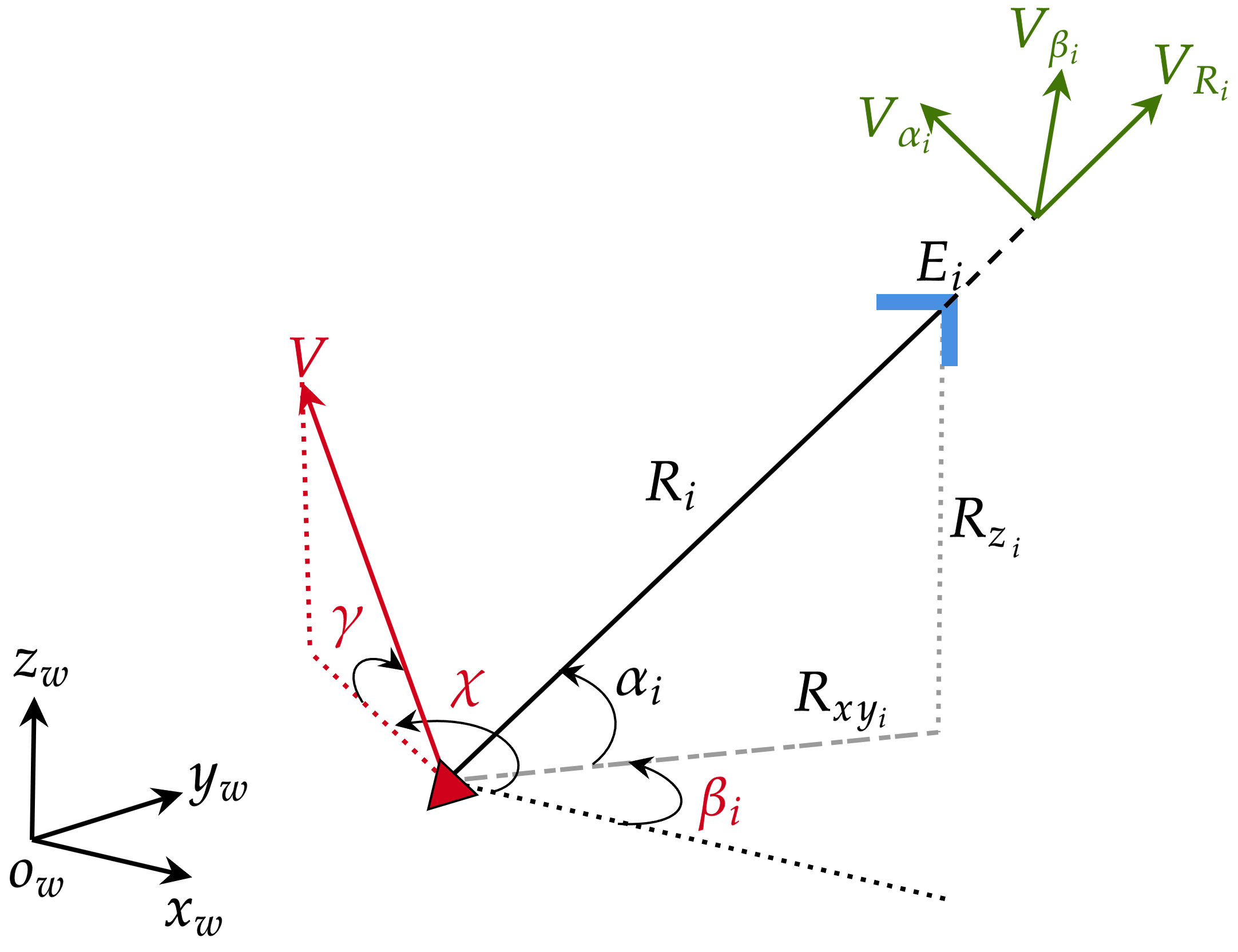}}
	\caption{Instantaneous relative geometry between quadrotor and $E_i$.}
	\label{relative}
\end{figure}

\begin{figure}[htbp]
	\centering
	\includegraphics[scale=0.1]{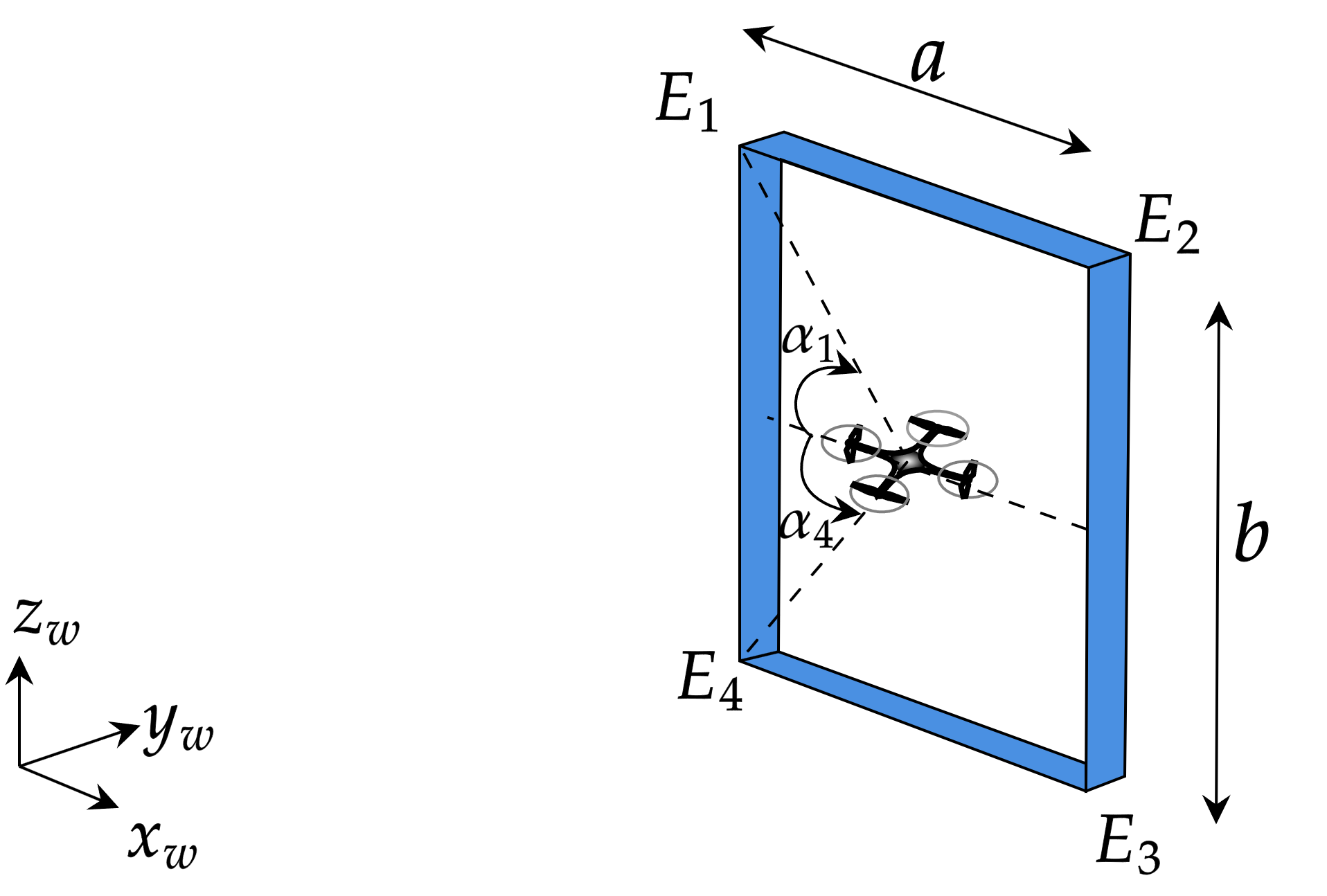}
	\caption{Quadrotor at centroid of the window.}
	\label{atcentroid}
\end{figure}

Consider an instantaneous relative geometry between the quadrotor and $i^\text{th}$ vertex of the window $E_i$, as illustrated in Fig. \ref{relative}. 
Let $R_i$ denote the line-of-sight distance between the quadrotor and $E_i$, with $R_{xy_i}$ and $R_{z_i}$ denoting its components in the $x_w-y_w$ plane, and along the $z_w$ axis, respectively. Let $\alpha_i$, and $\beta_i$ denote the elevation and azimuth angles of $E_i$ with respect to the vehicle. Accordingly,
\begin{align}\label{cos}
	\beta_i&=\tan^{-1}\left(\frac{y_i-y}{x_i-x}\right)\\ \alpha_i&=\sin^{-1}\left(\frac{R_{z_i}}{R_i}\right) \label{sin}
\end{align}
where $\beta_i\in[0,\pi]$ and $\alpha_i\in[-\frac{\pi}{2},\frac{\pi}{2}]$.
The line-of-sight separation $R_i$ is given by
\begin{equation}
	{R_i}=\left((x_i-x)^2+(y_i-y)^2+(z_i-z)^2\right)^{\frac{1}{2}}
\end{equation}
The relative velocity components of $i^{\text{th}}$ vertex of the window with respect to the quadrotor are denoted as $V_{R_i},V_{\alpha_i}$, and $V_{\beta_i}$. Here, $V_{R_i}$ represents the relative velocity along the line joining the quadrotor and $E_i$, while $\{V_{\alpha_i},V_{\beta_i}\}$ are the relative velocity components orthogonal to it, with $V_{\alpha_i}$ parallel to $x_w-y_w$ plane. The three-dimensional kinematic equations of relative motion of the $i^{\text{th}}$ vertex of the window with respect to the quadrotor can be expressed as \cite{Chakravarthy2012}
\begin{align}\label{3D:Kinematic1}
	V_{R_i} &= \dot{R}_i = -V\left( \cos{\gamma} \cos{\alpha_i} \cos(\chi - \beta_i) + \sin{\gamma} \sin{\alpha_i} \right) \\ 
	\label{3D:Kinematic2}
	V_{\alpha_i} &= R_i \dot{\alpha}_i = -V \left( -\cos{\gamma} \sin{\alpha_i} \cos(\chi - \beta_i) \right. \notag \\
	&\qquad\quad + \left. \sin{\gamma} \cos{\alpha_i} \right) \\ 
	\label{3D:Kinematic3}
	V_{\beta_i} &= R_i \dot{\beta}_i \cos{\alpha_i} = -V \cos{\gamma} \sin(\chi - \beta_i)
\end{align}
Considering the window with dimensions $E_1E_2=a$ and $E_1E_4=b$ as shown in Fig. \ref{atcentroid}, the desired azimuth and elevation angles when the vehicle is at the centroid of the window can be expressed using Eqs. \eqref{cos} and \eqref{sin} as
\begin{align}
	\beta_1(t_T) &= \beta_4(t_T) = 180 \text{ deg}, \beta_2(t_T) = \beta_3(t_T) = 0 \label{c2} \\
	\alpha_1(t_T) &= -\alpha_4(t_T) = \alpha_2(t_T) =\notag \\
	&\qquad -\alpha_3(t_T) =  \sin^{-1}\left(\frac{b}{\sqrt{a^2 + b^2}}\right) \label{c1}
\end{align}
where $t_T$ is the time at which the quadrotor reaches the centroid of the window. Further, the proposed guidance solution comprises two angular components; a desired flight path angle and a desired heading angle, which are described subsequently.
\subsection{Proposed flight path angle}
\begin{figure}[htbp]
	\centering
	\includegraphics[scale=0.06]{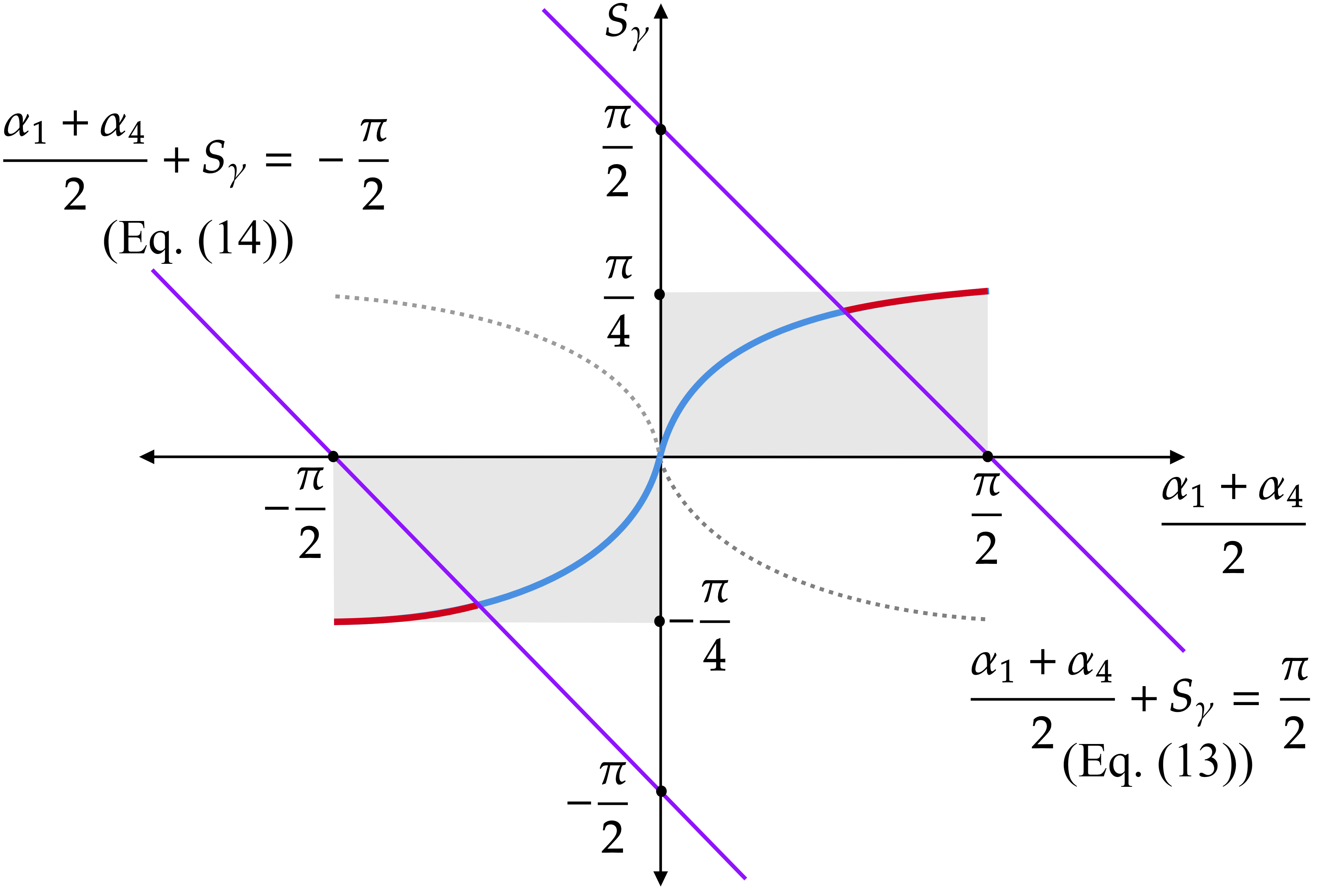}
	\caption{Elliptic shaping angle profile for proposed flight path angle, $S_\gamma$.}
	\label{ellipse1}
\end{figure}

The desired flight path angle is proposed using elevation angles of the vertices $E_1$ and $E_4$
\begin{subnumcases}{\label{desg}\gamma_{des} =}
	\frac{\alpha_1 + \alpha_4}{2} + S_\gamma \notag \\
	\qquad \quad \quad \forall\; \frac{\alpha_1 + \alpha_4}{2} + S_\gamma \in \left[-\frac{\pi}{2}, \frac{\pi}{2}\right] \label{desa} \\
	\pi - \left( \frac{\alpha_1 + \alpha_4}{2} + S_\gamma \right) \notag \\
	\qquad \quad \quad \forall\; \frac{\alpha_1 + \alpha_4}{2} + S_\gamma > \frac{\pi}{2} \label{desb} \\
	-\left( \pi + \frac{\alpha_1 + \alpha_4}{2} + S_\gamma \right) \notag \\
	\qquad \quad \quad \forall\; \frac{\alpha_1 + \alpha_4}{2} + S_\gamma < -\frac{\pi}{2} \label{desc}
\end{subnumcases}
where the term $\frac{\alpha_1+\alpha_4}{2}$ and $S_\gamma$ represent the angular bisector component and shaping angle component, respectively. 
Further, the shaping angle function $S_\gamma$ in Eq. \eqref{desg} is proposed as
\begin{subnumcases}{\label{des2}S_{\gamma}=}
	\sgn\left( \frac{\alpha_1 + \alpha_4}{2} \right)
	\frac{\left|\pi^2 - 4\left( \frac{\alpha_1 + \alpha_4}{2} + \frac{\pi}{2} \right)^2\right|^{1/2}}{4}, \notag\\
	\qquad\qquad \qquad\text{for } \frac{\alpha_1 + \alpha_4}{2} \in \left[-\frac{\pi}{2}, 0 \right]\label{des2a} \\
	\sgn\left( \frac{\alpha_1 + \alpha_4}{2} \right)
	\frac{\left|\pi^2 - 4\left( \frac{\alpha_1 + \alpha_4}{2} - \frac{\pi}{2} \right)^2\right|^{1/2}}{4}, \notag\\
\qquad\qquad \qquad \text{for } \frac{\alpha_1 + \alpha_4}{2} \in \left( 0, \frac{\pi}{2} \right]\label{des2b}
\end{subnumcases}
where $\sgn$ represents the signum function.
Figure \ref{ellipse1} illustrates the elliptical relationship between the shaping angle and the angular bisector components. 
The key idea behind the proposed guidance approach (Eqs. \eqref{desg} and \eqref{des2}) is to adjust the shaping angle as a function of the angular bisector to achieve the desired flight path angle for flying through the window centroid satisfying Eq. \eqref{c1}. The elliptical shaping angle profile illustrated in Fig. \ref{ellipse1} comprises three operating regions separated by two violet lines defined as
\begin{align}
	\frac{\alpha_1+\alpha_4}{2}+S_\gamma&=\frac{\pi}{2}\label{3D:L1}\\
	\frac{\alpha_1+\alpha_4}{2}+S_\gamma&=-\frac{\pi}{2}\label{3D:L2}
\end{align}
\begin{figure}[] 
	\centering
	\subfloat[{} \label{3D:M1}]{\includegraphics[ width=6.20cm]{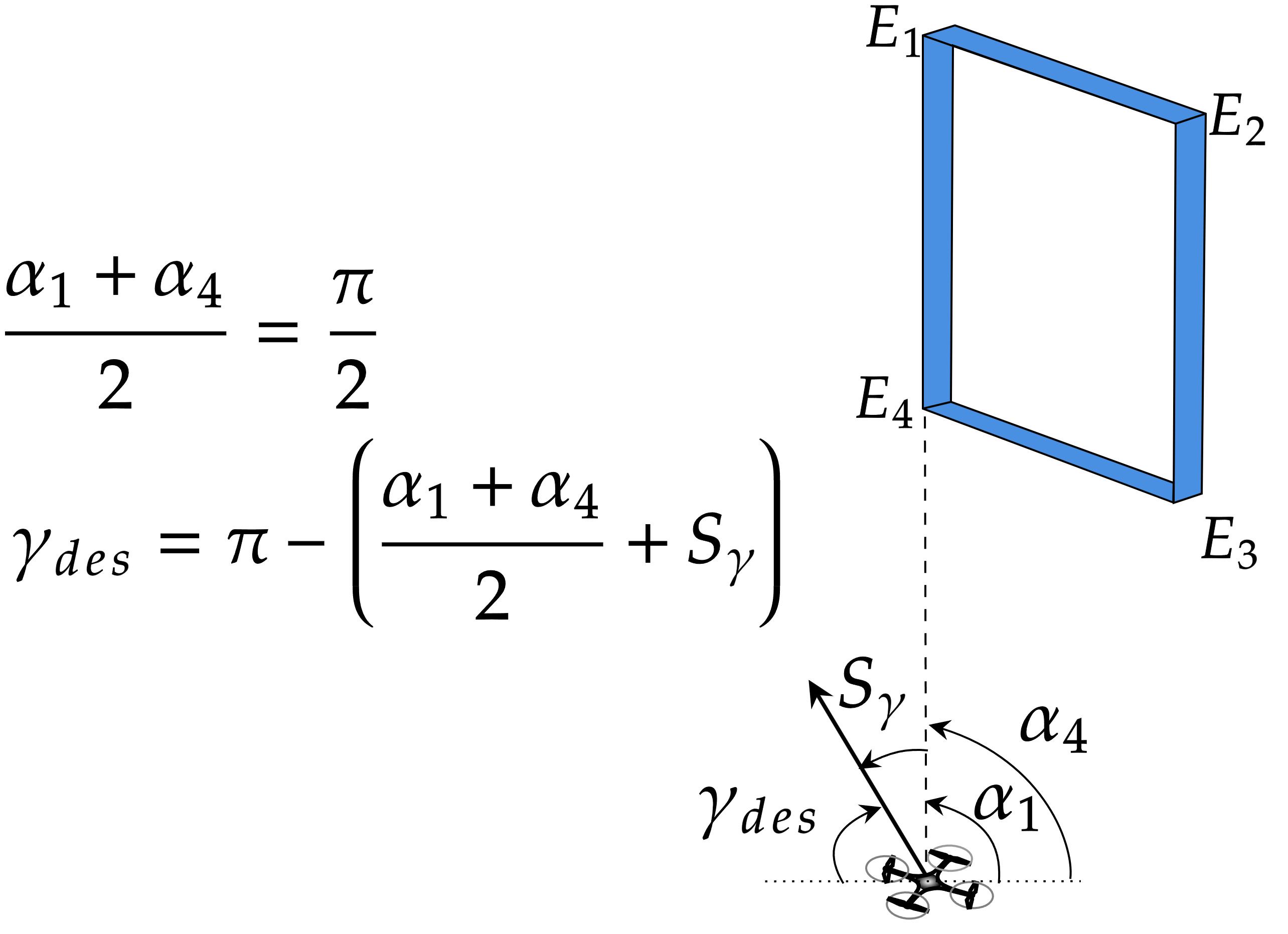}}\\
	\subfloat[{} \label{3D:M2}] {\includegraphics[width=6.20cm]{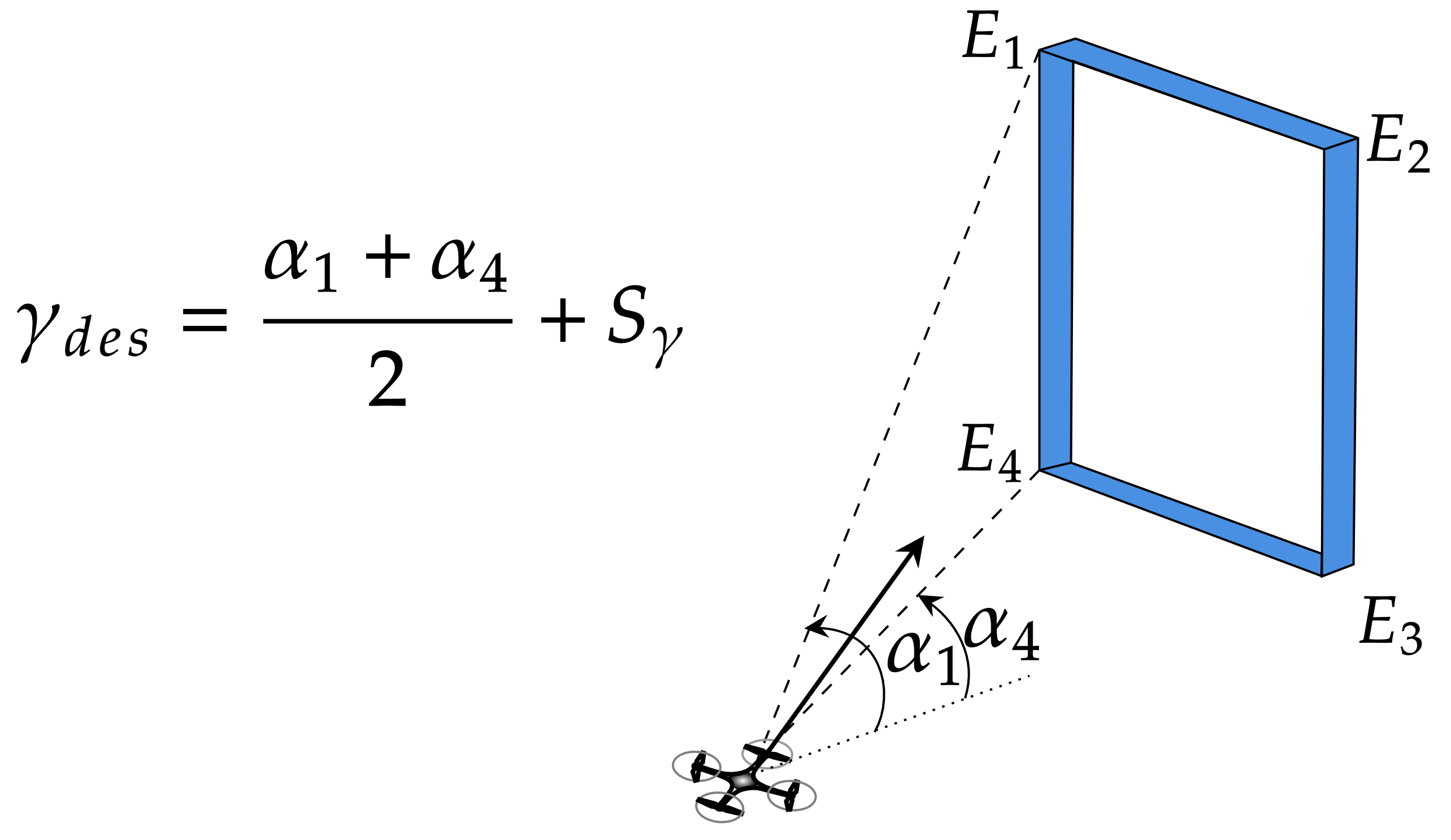}}\\
	\subfloat[{}   \label{3D:M3}]{\includegraphics[width=6.20cm]{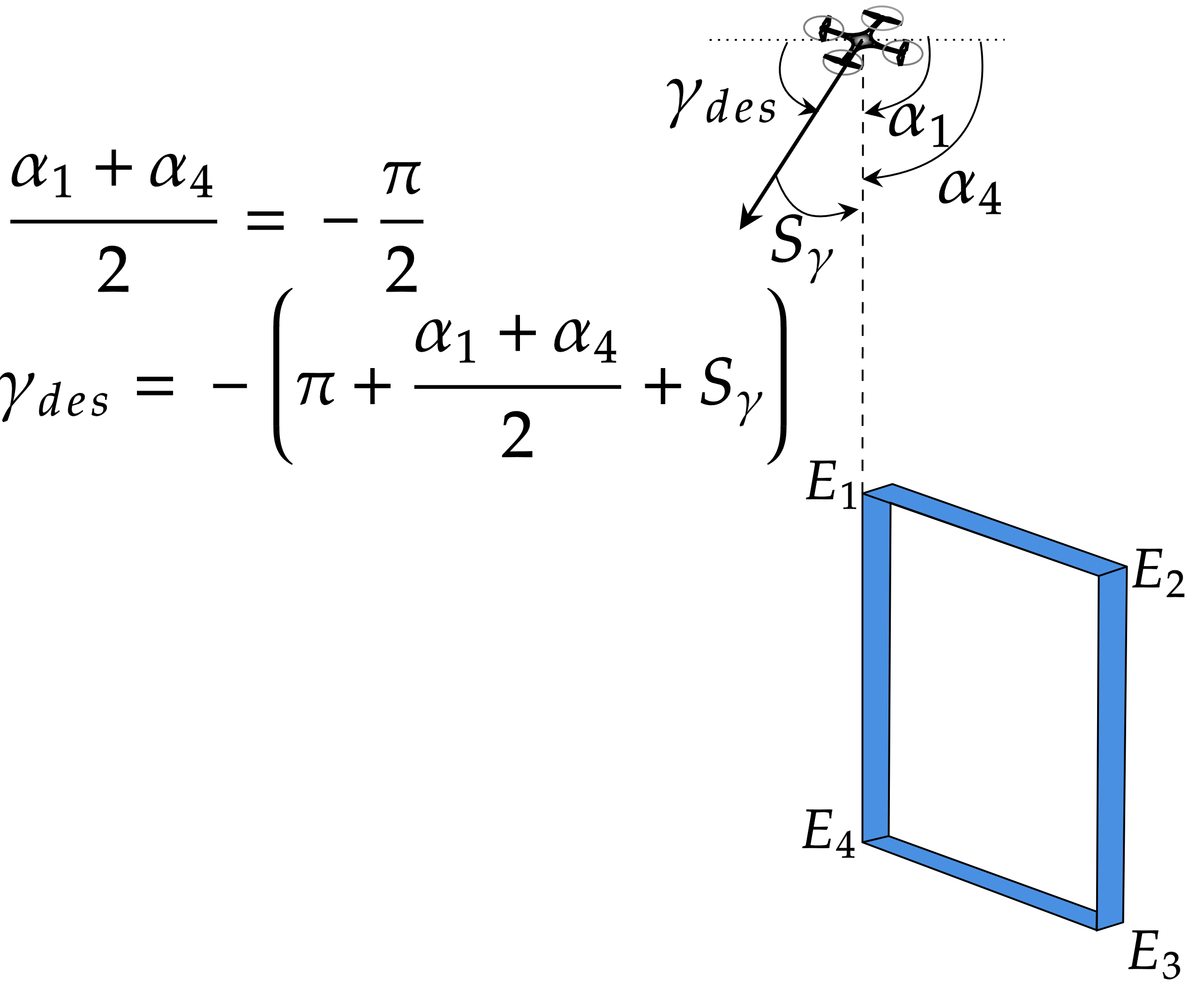}}
	\caption{Instantaneous desired flight path angle along the guided traversal trajectory. (a) Limiting geometry. (b) Intermediate point. (c) Limiting geometry.}
	\label{3D:M}
\end{figure}
The blue curve within the shaded region in Fig. \ref{ellipse1} contributes to the desired flight path angle as governed by Eq. \eqref{desa}. 
Equations \eqref{desb} and \eqref{desc} provide a framework for the proposed command to be within its permissible values, that is, $\gamma_{des}\left[-\frac{\pi}{2},\frac{\pi}{2}\right]$. 
The red curve to the right of the line defined by Eq. \eqref{3D:L1} determines the desired flight path angle according to Eq. \eqref{desb}. Therein, since $\frac{\alpha_1+\alpha_4}{2}+S_\gamma>\frac{\pi}{2}$, the desired flight path angle is constructed to lie in $\left(0,\frac{\pi}{2}\right)$. Similarly, the red curve to the left of the line defined by Eq. \eqref{3D:L2} determines the desired flight path angle as in Eq. \eqref{desc}, ensuring $\gamma_{des}\in \left(-\frac{\pi}{2},0\right)$ when $\frac{\alpha_1+\alpha_4}{2}+S_\gamma < -\frac{\pi}{2}$. Considering some representative engagement geometries, we further discuss the geometric understanding of the proposed method.

Figure \ref{3D:M} considers three specific instantaneous geometries, displaying corresponding shaping angle values and desired flight path angles as governed by Eqs. \eqref{des2} and \eqref{desg}, respectively.
Consider an instantaneous limiting geometry where the quadrotor makes elevation angles of $\alpha_1=\alpha_4=\frac{\pi}{2}$ rad, as shown in Fig. \ref{3D:M1}. From Eq. \eqref{des2} and as illustrated in Fig. \ref{ellipse1}, the shaping angle requirement $S_\gamma$ for $\frac{\alpha_1+\alpha_4}{2}=\frac{\pi}{2}$ is $\frac{\pi}{4}$. This implies $\frac{\alpha_1+\alpha_4}{2}+S_\gamma=\frac{3\pi}{4}>\frac{\pi}{2}$, which corresponds to a point on the right red curve in Fig. \ref{ellipse1}. With this, the guidance command in  Eq. \eqref{desb} commands the quadrotor with a flight path angle of $\gamma_{des}=\frac{\pi}{4}$. Using $\gamma_{des}=\frac{\pi}{4}$, the quadrotor attains necessary lateral clearance from window plane before traversing through the centroid. Next, consider an intermediate geometry as shown in Fig. \ref{3D:M2}, where $
0<\frac{\alpha_1+\alpha_4}{2}<\frac{\pi}{2}$. According to \eqref{des2b}, a positive shaping angle is added to the angular bisector component to shape the desired flight path angle. This corresponds to the blue curve shown in Fig. \ref{ellipse1} and it can be noted that,
as the bisector component approaches zero, the magnitude of the shaping angle also decreases to zero and with the bisector pointing towards the centroid, the quadrotor eventually traverses through it.
Figure \ref{3D:M3} illustrates another limiting scenario similar to the one shown in \ref{3D:M1} with $\alpha_1=\alpha_4=-\frac{\pi}{2}=\frac{\alpha_1+\alpha_4}{2}=-\frac{\pi}{2}$. Here, Eq. \eqref{desc} generates $\gamma_{des}=-\frac{\pi}{4}$, which provides the necessary lateral clearance for traversing through the window centroid.
In order to achieve three-dimensional traversal, along with the proposed flight path angle, a desired heading angle needs to be designed, which is discussed subsequently.
\subsection{Proposed heading angle} 
Utilizing the azimuth angles $\beta_1$ and $\beta_2$, the desired heading angle is proposed as
\begin{subnumcases}{\label{desheading}\chi_{des} =}
	\frac{\beta_1+\beta_2}{2}+S_{\chi} \;  \forall \; \frac{\alpha_1+\alpha_4}{2}+S_\gamma\in\left[-\frac{\pi}{2},\frac{\pi}{2}\right]\label{desheadingnew}\\
	-\left(\frac{\beta_1+\beta_2}{2}+S_{\chi}\right) \quad \text{otherwise}\label{desheadingother}
\end{subnumcases}
where the terms $\frac{\beta_1+\beta_2}{2}$ and $S_{\chi}$ represent the angular bisector component and the shaping angle component, respectively.
\begin{figure}[htbp]
	\centering
	\includegraphics[scale=0.065]{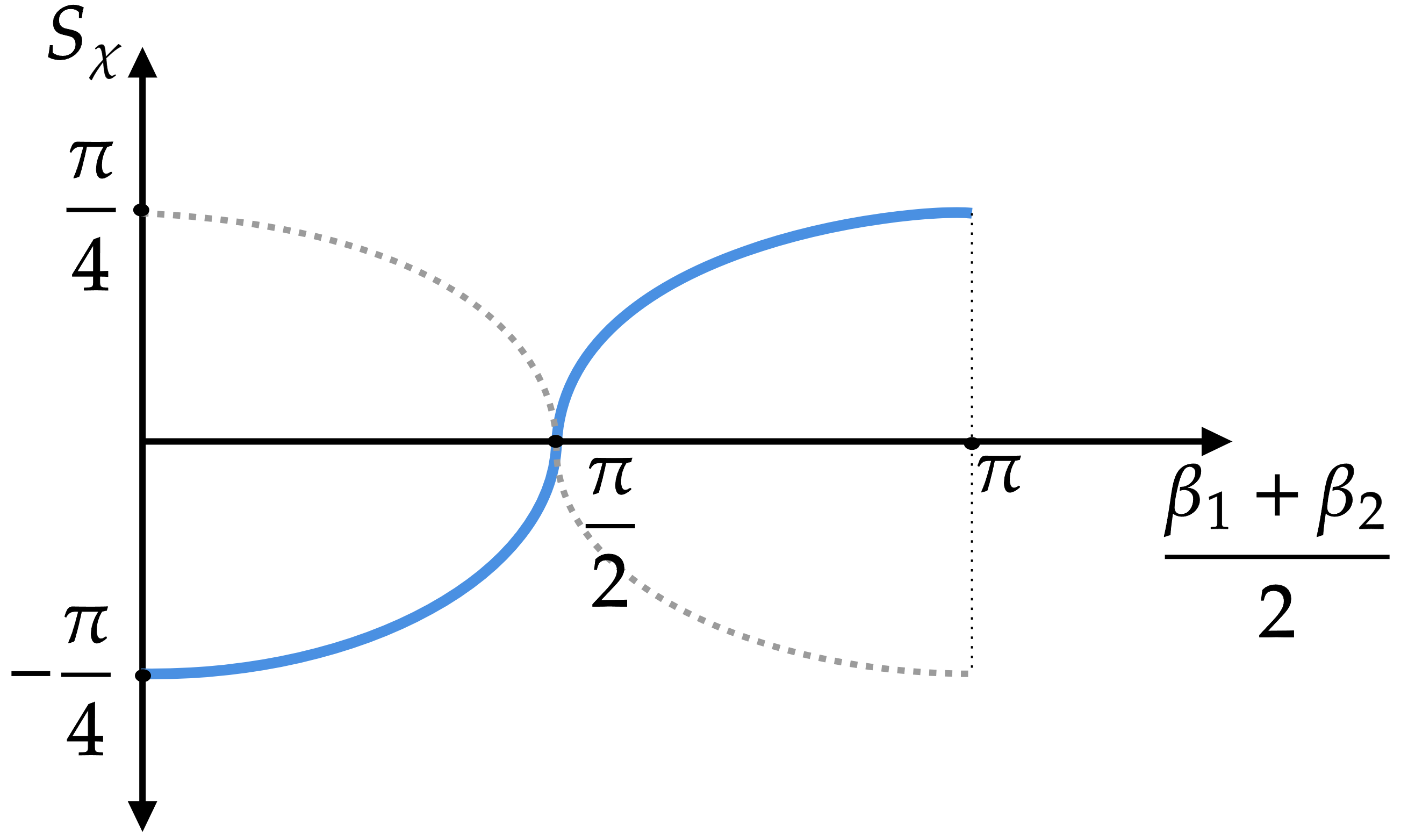}
	\caption{Elliptic shaping angle profile for proposed heading angle, $S_\chi$.}
	\label{ellipse2}
\end{figure} 
The shaping angle component in Eq. \eqref{desheading} is governed by the elliptical relationship, 
\begin{align}{\label{desheading2}
		S_\chi=}
	\sgn\left(\frac{\beta_1+\beta_2}{2}-\frac{\pi}{2}\right)\frac{\big|\pi^2-(\beta_1+\beta_2)^2\big|^{1/2}}{4}, \notag\\
	\qquad\qquad \qquad \text{for $\beta_1(t)$, $\beta_2(t) \in [0,\pi]$ }
\end{align}
The shaping angle profile is plotted against angular bisector component in Fig. \ref{ellipse2}.
For the geometries where $\frac{\alpha_1+\alpha_4}{2}+S_\gamma$ falls within $\left[-\frac{\pi}{2},\frac{\pi}{2}\right]$, the logic behind the proposed heading angle in Eq. \eqref{desheadingnew} is similar to the one presented in \cite{mekjrnl}. 
The shaping angle $S_\chi$ as governed by Eq. \eqref{desheading2} and as illustrated in Fig. \ref{ellipse2} is adjusted in accordance with the angular bisector component $\frac{\beta_1+\beta_2}{2}$ to direct the vehicle towards the centroid of the window. From Eq. \eqref{desheading2}, the shaping angle varies monotonically to zero as the bisector component $\frac{\beta_1+\beta_2}{2}$ approaches $\frac{\pi}{2}$, which using Eq. \eqref{desheadingnew} leads to the desired entry heading angle of $\frac{\pi}{2}$ at the centroid. For geometries which corresponds to $\frac{\alpha_1+\alpha_4}{2}+S_\gamma\notin\left[-\frac{\pi}{2},\frac{\pi}{2}\right]$, that is, the ones characterized by Eqs. \eqref{desb} and \eqref{desc}, the quadrotor heading direction is evaluated using Eq. \eqref{desheadingother}. In conjuction with Eqs. \eqref{desb} and \eqref{desc}, the relation in Eq. \eqref{desheadingother} captures the effect of changes required for maintaining $\gamma_{des}\in\left[-\frac{\pi}{2},\frac{\pi}{2}\right]$. 
\begin{remark}	The proposed flight path angle in Eq. \eqref{desg} and shaping angle in Eq. \eqref{des2} are functions of elevation angles relative to vertices $E_2$ and $E_3$. In a similar way, these angles can also be expressed using the elevation angles of the vertices $E_2$ and $E_3$.
	 Similarly, the proposed heading angle in Eq. \eqref{desheading}, and the corresponding shaping function in Eq. \eqref{desheading2} can also be formulated using azimuth angles relative to vertices $E_3$ and $E_4$.
\end{remark}
\section{STABILITY CHARACTERISTICS AND PHASE PORTRAIT ANALYSIS}\label{stability}
This section discusses the stability characteristics of the proposed guidance strategy, followed by phase portrait analysis in the bearing angle space. 
\begin{theorem}
	Governed by the system Eqs. \eqref{pointmass}-\eqref{pointmassend}, \eqref{3D:Kinematic1}-\eqref{3D:Kinematic3}, \eqref{desg}, \eqref{des2}, \eqref{desheading}, and \eqref{desheading2}, the quadrotor motion is asymptotically stable about the line passing through the centroid of the window and normal to the window plane.
\end{theorem} 

\begin{proof}
	Figure \ref{3D:stability} shows an instantaneous geometry of a quadrotor approaching the window. 
Let $D$ be the instantaneous distance of the quadrotor from the line normal to the window plane and passing through the centroid.
Further, let $D_{x}$ and $D_z$ represent the horizontal and vertical displacements of the quadrotor with respect to that line, respectively, as shown in Fig. \ref{3D:stability}.
	Using the geometries depicted in Figs. \ref{relative} and \ref{3D:stability} the following  relations can be readily deduced, 
	\begin{align}\label{3D:D_x}
		D_{x}&=-\left(R_1 \cos\alpha_1 \cos\beta_1+\frac{a}{2}\right)\\
		D_z&=-\left(R_1\sin{\alpha_1}-\frac{b}{2}\right)\label{3D:D_z}
	\end{align}
	From Figs. \ref{relative} and \ref{3D:stability}, the window dimensions $a$ and $b$ can be expressed in terms of line-of-sight distances and bearing angles with respect to vertices $E_1$, $E_2$, and $E_4$ as 
	\begin{align}\label{3D:a}
		a&=R_2\cos\alpha_2\cos\beta_2-R_1\cos\alpha_1\cos\beta_1\\
		b&=R_1\sin \alpha_1-R_4\sin\alpha_4\label{3D:b}
	\end{align}
	Using Eqs \eqref{3D:a} and \eqref{3D:b} in Eqs. \eqref{3D:D_x} and \eqref{3D:D_z} yields
	\begin{align}\label{3D:D_X}
		D_{x}&=-\frac{1}{2}\left(R_1 \cos\alpha_1 \cos\beta_1+R_2 \cos\alpha_2 \cos\beta_2\right)\\
		D_{z}&=-\frac{1}{2}\left(R_1\sin \alpha_1+R_4\sin\alpha_4 \label{3D:D_Z}  \right)
	\end{align}
	Consider a Lyapunov function,
	\begin{equation}\label{W}
		W=\frac{1}{2}\left(D^2_{x}+D^2_z\right)
	\end{equation}
	Here, $W\geq0$ for all $(D_{x},D_{z})\in \Bbb {R}^2$ with $W=0$ if and only if $D_{x}=D_{z}=0$. Hence the Lyapunov function $W$ is positive definite in the domain $\Bbb {R}^2$.  
	\begin{figure}[]
		\centering
		\includegraphics[scale=0.0635]{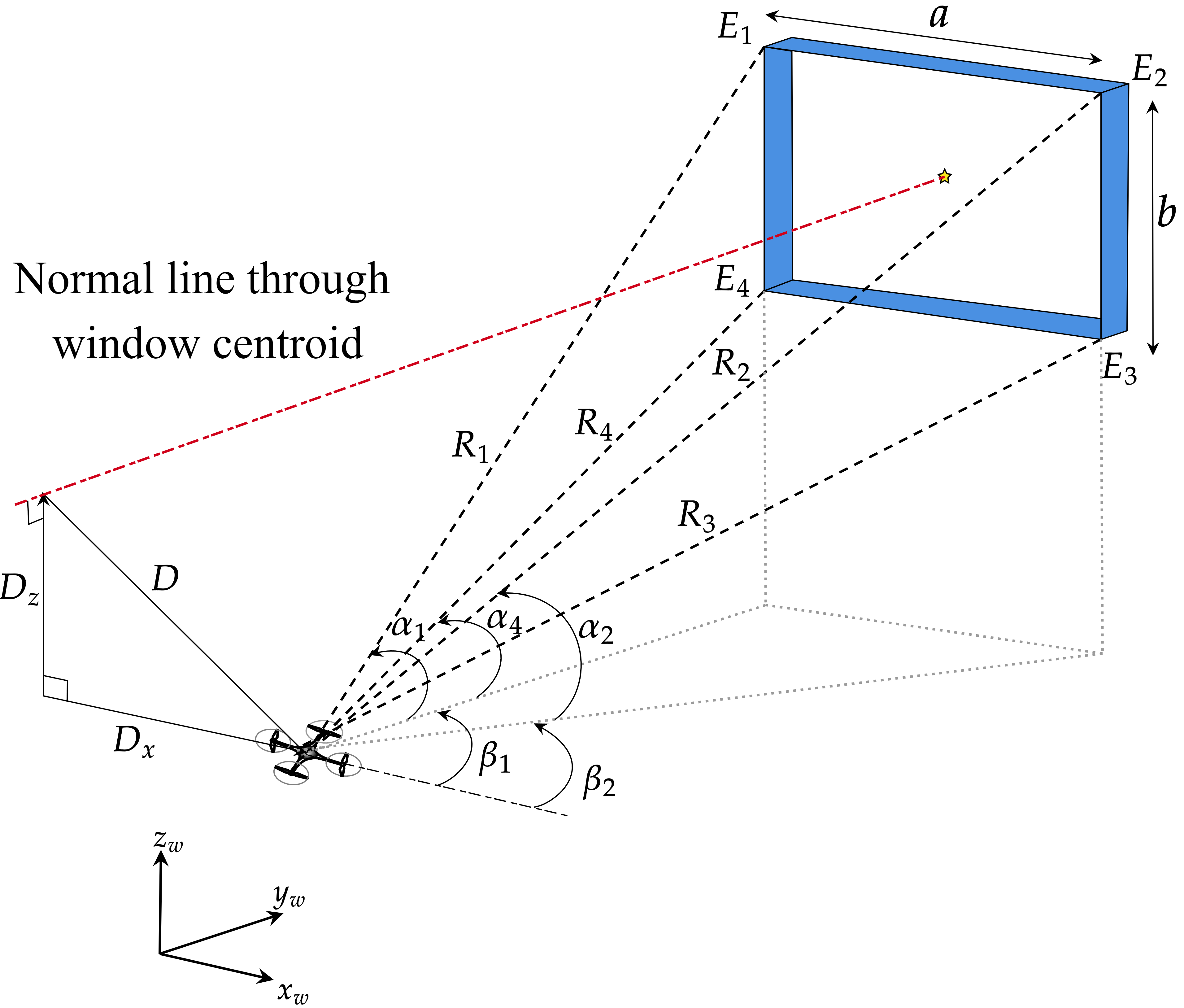}
		\caption{Instantaneous geometry of quadrotor moving towards the window.}
		\label{3D:stability}
	\end{figure}
	Taking the time derivative of the Lyapunov function in Eq. \eqref{W},
	\begin{equation}\label{3D:Lypfunction}
		\dot{W}=D_{x}\dot{D}_{x}+D_z\dot{D}_z
	\end{equation}
	Differentiating Eq. \eqref{3D:D_X} with respect to time,
\begin{align}\label{3D:dotDx}
	\dot{D}_x = \frac{1}{2}\Biggr( 
	&-R_1\dot{\beta}_1\cos\alpha_1\sin\beta_1 \notag\\
	&+ \cos\beta_1 \left( -R_1\dot{\alpha}_1\sin\alpha_1 + \dot{R}_1\cos\alpha_1 \right) \notag\\
	&-R_2\dot{\beta}_2\cos\alpha_2\sin\beta_2 \notag\\
	&+ \cos\beta_2 \left( -R_2\dot{\alpha}_2\sin\alpha_2 + \dot{R}_2\cos\alpha_2 \right)
	\Biggr)
\end{align}
	Using Eqs. \eqref{3D:Kinematic1}-\eqref{3D:Kinematic3}, $\gamma=\gamma_{des}$, and $\chi=\chi_{des}$ in Eq. \eqref{3D:dotDx},
	\begin{equation}\label{3D:Dx_final}
		\dot{D}_x=V\cos\gamma_{des}\cos{\chi_{des}}
	\end{equation}
	Differentiating Eq. \eqref{3D:D_Z} with respect to time,
\begin{align}\label{3D:DZ_Dot}
	\dot{D}_z = \frac{1}{2} \Big( R_1\dot{\alpha}_1\cos{\alpha_1} + \dot{R}_1\sin{\alpha_1} 
	\notag \\
	+ R_4\dot{\alpha}_3\cos{\alpha_4} + \dot{R}_3\sin{\alpha_4} \Big)
\end{align}
	Using Eqs. \eqref{3D:Kinematic1}-\eqref{3D:Kinematic3} and $\gamma=\gamma_{des}$ in Eq. \eqref{3D:DZ_Dot},
	\begin{equation}\label{3D:Dzdot_final}
		\dot{D}_z=V\sin\gamma_{des}
	\end{equation}
	Using Eqs. \eqref{3D:Dx_final} and \eqref{3D:Dzdot_final} in  Eq. \eqref{3D:Lypfunction},
	\begin{equation}\label{3D:Final LyaD}			\dot{W}=D_xV\cos\gamma_{des}\cos{\chi_{des}}+D_zV\sin\gamma_{des}
	\end{equation}
	From the geometry shown in Fig. \ref{3D:stability}, it can be noted that
	\begin{align}\label{3D:SimilarT1}
		\frac{\cos\alpha_1}{\cos\alpha_4}&=\frac{R_4}{R_1}\\
		\frac{R_1\cos\alpha_1}{R_2\cos\alpha_2}&=\frac{\sin\beta_2}{\sin\beta_1}\label{3D:SimilarT2}
	\end{align} 
	Rest of the analysis is divided into four quadrants around the centroid of the window.
	\begin{enumerate}
		\item Case 1: $D_z<0$ and $D_x<0$\newline
	From the geometry shown in Fig. \ref{3D:stability}, it can be noted that
		\begin{equation}\label{dzrangealpha}
			\alpha_1,\alpha_2\in\left(0,\frac{\pi}{2}\right)\quad \forall \; D_z<0
	\end{equation}
		From Eq. \eqref{3D:D_Z}, when $D_z<0$
		\begin{equation}
			R_1\sin\alpha_1+R_4\sin\alpha_4>0
		\end{equation}
		which implies
		\begin{equation}\label{3D:relation}
			\frac{R_1}{R_4}\sin\alpha_1+\sin\alpha_4>0
		\end{equation}
	Using  Eq. \eqref{3D:SimilarT1} in Eq. \eqref{3D:relation} while utilizing Eq. \eqref{dzrangealpha}, 
		\begin{eqnarray}
			\implies\frac{\cos\alpha_4}{\cos\alpha_1}\sin\alpha_1+\sin\alpha_4>0\nonumber\\
			\implies \sin(\alpha_1+\alpha_4)>0\nonumber\\	
		\implies	\frac{\alpha_1+\alpha_4}{2}\in\left(0,\frac{\pi}{2}\right)\quad \forall \; D_z<0 	\label{3D:bisector1}
		\end{eqnarray}
	From the proposed guidance command in Eqs. \eqref{desa} and \eqref{desb} with shaping angle in Eq. \eqref{des2b}, it can be deduced that
		\begin{equation}
			\gamma_{des}\in\left(0,\frac{\pi}{2}\right)\quad \forall \;\frac{\alpha_1+\alpha_4}{2}\in\left(0,\frac{\pi}{2}\right)\label{3D:gammaL}
		\end{equation}
	Further, from Eq. \eqref{3D:D_X}, when $D_x<0$ 
		\begin{equation}
			R_1 \cos\alpha_1 \cos\beta_1+R_2 \cos\alpha_2 \cos\beta_2>0
		\end{equation}
		which implies
		\begin{equation}\label{3D:relation2}
			\frac{R_1\cos\alpha_1}{R_2\cos\alpha_2}\cos\beta_1+\cos\beta_2>0
		\end{equation}
		Using Eq. \eqref{3D:SimilarT2} in Eq. \eqref{3D:relation2} with $D_x<0$,
		\begin{eqnarray}
			\implies\frac{\sin\beta_2}{\sin\beta_1}\cos\beta_1+\cos\beta_2>0\nonumber\\
			%\implies \sin\beta_2\cos\beta_1+\sin\beta_1\cos\beta_1>0\\
			\implies \sin(\beta_1+\beta_2) >0\nonumber\\
		\implies	\frac{\beta_1+\beta_2}{2}\in\left(0,\frac{\pi}{2}\right) \quad \forall\; D_x<0 \label{3D:bisector4}
		\end{eqnarray}
	From the proposed guidance command in Eq. \eqref{desheading} with shaping angle in Eq. \eqref{desheading2}, it can be deduced that
		\begin{align}\label{3D:chiL}
			\chi_{des}&\in\left(-\frac{\pi}{2},\frac{\pi}{2}\right) \quad \forall \; \frac{\beta_1+\beta_2}{2}\in\left(0,\frac{\pi}{2}\right)
		\end{align}
		Using Eqs. \eqref{3D:gammaL} and \eqref{3D:chiL} in Eq. \eqref{3D:Final LyaD}
		\begin{equation}\label{3D:1}
			\dot{W}<0 \quad \forall \; D_z<0, D_x<0
		\end{equation}
		\item Case 2: $D_z>0$ and $D_x>0$\newline
	From the geometry shown in Fig. \ref{3D:stability}, it can be noted that
			\begin{equation}\label{newlyadded}
				\alpha_1,\alpha_2\in\left(-\frac{\pi}{2},0\right)\quad \forall \; D_z>0
		\end{equation}
		From Eq. \eqref{3D:D_Z}, when $D_z>0$
		\begin{equation}
			R_1\sin\alpha_1+R_4\sin\alpha_4<0
		\end{equation}
		which implies
		\begin{equation}\label{3D:relation1}
			\frac{R_1}{R_4}\sin\alpha_1+\sin\alpha_4<0
		\end{equation}
		Using  Eq. \eqref{3D:SimilarT1} in Eq. \eqref{3D:relation1} while utilizing \eqref{newlyadded},
		\begin{eqnarray}
			\implies\frac{\cos\alpha_4}{\cos\alpha_1}\sin\alpha_1+\sin\alpha_4<0 \nonumber\\
		%	\implies \cos\alpha_4\sin\alpha_1+\sin\alpha_4\cos\alpha_1<0\\
			\implies \sin(\alpha_1+\alpha_4)<0\nonumber\\
			\implies \frac{\alpha_1+\alpha_4}{2}\in\left(-\frac{\pi}{2},0\right)\quad \forall \; D_z>0 \label{3D:bisector3}
		\end{eqnarray}
From the proposed guidance command Eqs. \eqref{desa} and \eqref{desc} with shaping angle in Eq. \eqref{des2a}, it can be deduced that	
		\begin{equation}
			\gamma_{des}\in\left(-\frac{\pi}{2},0\right)\quad \forall \;\frac{\alpha_1+\alpha_4}{2}\in\left(-\frac{\pi}{2},0\right)\label{3D:gammaL2}
		\end{equation}
		Further, from Eq. \eqref{3D:D_X}, when $D_x>0$ 
		\begin{equation}
			R_1 \cos\alpha_1 \cos\beta_1+R_2 \cos\alpha_2 \cos\beta_2<0
		\end{equation}
		which implies
		\begin{equation}\label{3D:relation3}
			\frac{R_1\cos\alpha_1}{R_2\cos\alpha_2}\cos\beta_1+\cos\beta_2<0
		\end{equation}
		Using Eq. \eqref{3D:SimilarT2} in Eq. \eqref{3D:relation3} with $D_x>0$,
		\begin{eqnarray}
			\implies\frac{\sin\beta_2}{\sin\beta_1}\cos\beta_1+\cos\beta_2<0\nonumber\\
			%\implies \sin\beta_2\cos\beta_1+\sin\beta_1\cos\beta_1<0\\
			\implies \sin(\beta_1+\beta_2) <0\nonumber\\
			\implies 
			\frac{\beta_1+\beta_2}{2}\in\left(\frac{\pi}{2},\pi\right) \quad \forall\; D_x>0 \label{3D:bisector2}
		\end{eqnarray}
From the proposed guidance command Eq. \eqref{desheading} with shaping angle in Eq. \eqref{desheading2}, it can be deduced that
		\begin{equation}\label{3D:chiL2}
			\chi_{des}\in\left(\frac{\pi}{2},\pi\right]\cup\left(-\pi,-\frac{\pi}{2}\right) \; \forall \; \frac{\beta_1+\beta_2}{2}\in\left(\frac{\pi}{2},\pi\right)
		\end{equation}
		Using Eqs. \eqref{3D:gammaL2} and \eqref{3D:chiL2} in Eq. \eqref{3D:Final LyaD}
		\begin{equation}\label{3D:2}
			\dot{W}<0 \quad \forall \; D_z>0, D_x>0
		\end{equation}
		\item Case 3: $D_z<0$ and $D_x>0$\newline
	Using Eqs. \eqref{3D:bisector1} and \eqref{3D:bisector2} in proposed guidance command Eqs. \eqref{desa}, \eqref{desb} and \eqref{desheading} with shaping angles Eq. \eqref{des2a} and Eq. \eqref{desheading2}, it can be deduced that
		\begin{align}
			\gamma_{des}&\in\left(0,\frac{\pi}{2}\right)\quad \forall \;\frac{\alpha_1+\alpha_4}{2}\in\left(0,\frac{\pi}{2}\right)\label{3D:gammaL3}\\
			\chi_{des}&\in\left(\frac{\pi}{2},\pi\right]\cup\left(-\pi,-\frac{\pi}{2}\right) \; \forall \; \frac{\beta_1+\beta_2}{2}\in\left(\frac{\pi}{2},\pi\right)	\label{3D:chiL3}
		\end{align}
		Using Eqs. \eqref{3D:gammaL3} and \eqref{3D:chiL3} in Eq. \eqref{3D:Final LyaD}
		\begin{equation}\label{3D:3}
			\dot{W}<0 \quad \forall \; D_z<0, D_x>0
		\end{equation}
		\item Case 4: $D_z>0$ and $D_x<0$ \newline
		Using Eqs. \eqref{3D:bisector4} and \eqref{3D:bisector3} in proposed guidance command 
			Eqs. \eqref{desa}, \eqref{desc} and \eqref{desheading} with shaping angles in Eqs. \eqref{des2a} and \eqref{desheading2}, it can be deduced that
		\begin{align}
			\gamma_{des}&\in\left(-\frac{\pi}{2},0\right)\quad \forall \;\frac{\alpha_1+\alpha_4}{2}\in\left(-\frac{\pi}{2},0\right)\label{3D:gammaL4}\\
			\label{3D:chiL4}
			\chi_{des}&\in\left(-\frac{\pi}{2},\frac{\pi}{2}\right) \quad \forall \; \frac{\beta_1+\beta_2}{2}\in\left(0,\frac{\pi}{2}\right)
		\end{align}
		Using Eqs. \eqref{3D:chiL4} and \eqref{3D:gammaL4} in Eq. \eqref{3D:Final LyaD},
		\begin{equation}\label{3D:4}
			\dot{W}<0 \quad \forall \; D_z>0, D_x<0
		\end{equation}
	\end{enumerate}
	From Eqs. \eqref{3D:Final LyaD}, \eqref{3D:1}, \eqref{3D:2}, \eqref{3D:3} and \eqref{3D:4}, it can be deduced that 
	\begin{align}
		\dot{W}&\leq 0 \quad \forall \;  (D_{x},D_{z})\in \Bbb {R}^2\\
		\dot{W}&=0\quad \text{at} \; D_{x}=0,D_{z}=0
	\end{align}
\end{proof}
\begin{figure}[htbp] 
	%\centering
	\subfloat[{} \label{ppgamma1}]
	{\includegraphics[width=8.75cm,trim={0.40cm 0.078cm 0.65cm 0.95cm},clip]{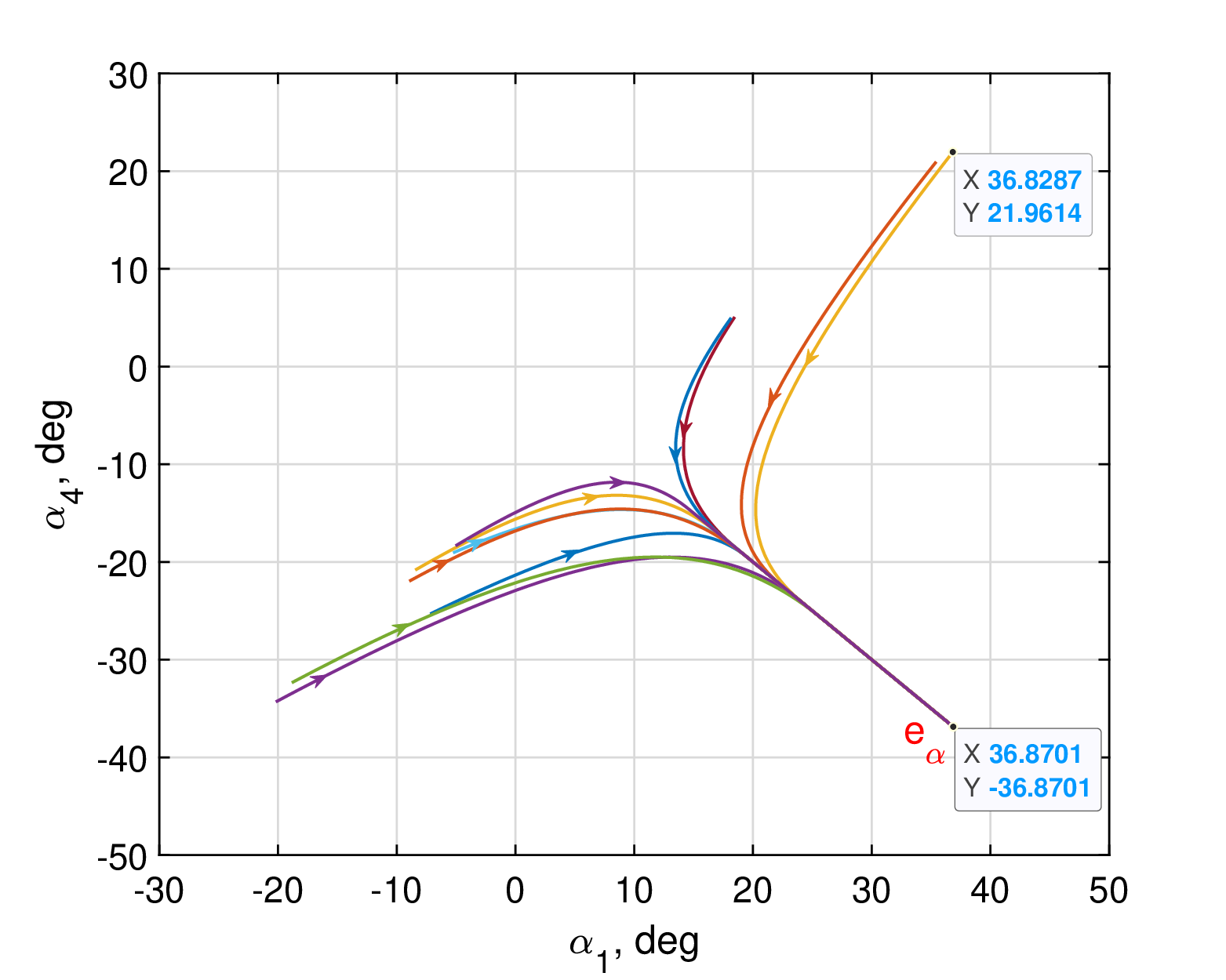}}\\
	\subfloat[{}  \label{ppbeta1}]
	{\includegraphics[width=8.75cm,trim={0.40cm 0.08cm 0.65cm 0.95cm},clip]{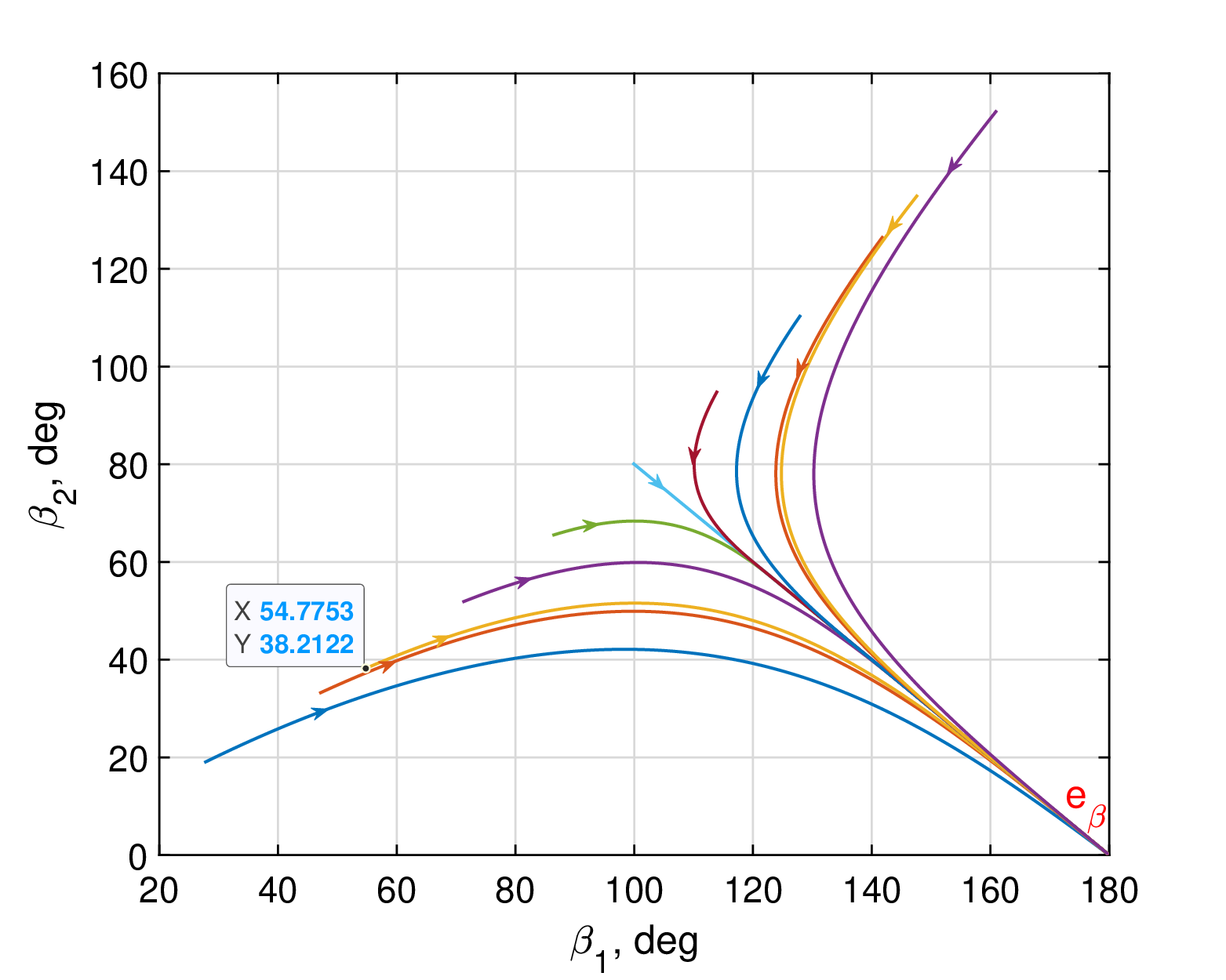}}
	\caption{Phase plane analysis. (a) Phase portraits in $\alpha_1-\alpha_4$ plane. (b) Phase portraits in $\beta_1-\beta_2$ plane.}
	\label{PP}
\end{figure}
We further analyze the behavior of the system described by the Eqs. \eqref{pointmass}-\eqref{pointmassend} under the proposed guidance commands given by Eqs. \eqref{desg} and \eqref{desheading}, where $\chi$ and $\gamma$ take their desired values. The analysis is carried out for a window of dimension of $a=4$ m and $b=3$ m. 
Considering the system states as $\alpha_1$, $\alpha_4$, $\beta_1$, and $\beta_2$, phase portraits are generated by numerically solving system Eqs. \eqref{pointmass}-\eqref{pointmassend} for 11 initial conditions with $\alpha_1(t_i),\alpha_4(t_i)\in[-\frac{\pi}{2},\frac{\pi}{2}]$ and  $\beta_1(t_i),\beta_2(t_i)\in[0,\pi]$. The resulting phase portraits are depicted in Fig. \ref{PP}.
From the window dimensions, and Eqs. \eqref{c2} and \eqref{c1}, the desired bearing angles from window centroid can be determined as $(\beta_1(t_T),\beta_2(t_T))=(180,0)$ deg and $(\alpha_1(t_T), \alpha_4(t_T)) = (36.87, -36.87)$ deg. Consider an initial condition where the bearing angles are $(\alpha_1(t_i), \alpha_4(t_i)) = (36.82, 21.96)$ deg and $(\beta_1(t_i), \beta_2(t_i)) = (54.77,38.21)$ deg. For this engagement geometry, the quadrotor employs the shaping angle component in Eq. \eqref{des2b} and the desired flight path angle in Eq. \eqref{desa} to maneuver from  $(\alpha_1(t_i), \alpha_4(t_i))$ to the final states $(\alpha_1(t_T), \alpha_4(t_T))$, as illustrated in Fig. \ref{ppgamma1}. The analysis shows that, starting from all initial conditions, the system states $(\alpha_1(t),\alpha_4(t))$ reach to their desired final point of $e_\alpha=(36.87,-36.87)$ deg by employing the proposed guidance commands, as illustrated in Fig. \ref{ppgamma1}. 
%The elevation angles reach their desired values of $\alpha_1(t_T)=-\alpha_4(t_T)=36.87$ deg, abiding by Eq. \eqref{c1}, indicating a successful window traversal. 
The corresponding $\beta_1-\beta_2$ phase portraits is shown in Fig. \ref{ppbeta1}. With the initial state $(\beta_1(t_i), \beta_2(t_i)) = (54.77,38.21)$ deg, the quadrotor utilizes the shaping angle component in Eq. \eqref{desheading2} and the desired heading angle in Eq. \eqref{desheadingnew} to achieve the desired traversal state $e_\beta = (180, 0)$ deg, as illustrated in Fig. \ref{ppbeta1}. All the trajectories in $\beta_1-\beta_2$ plane converge to the desired point $e_\beta$, as in Eq. \eqref{c2}, indicating the final desired state of traversal.
	\section{DYNAMIC MODEL AND CONTROL}\label{dynamics}
\begin{figure}[htbp]
		\centerline{\includegraphics[width=57mm]{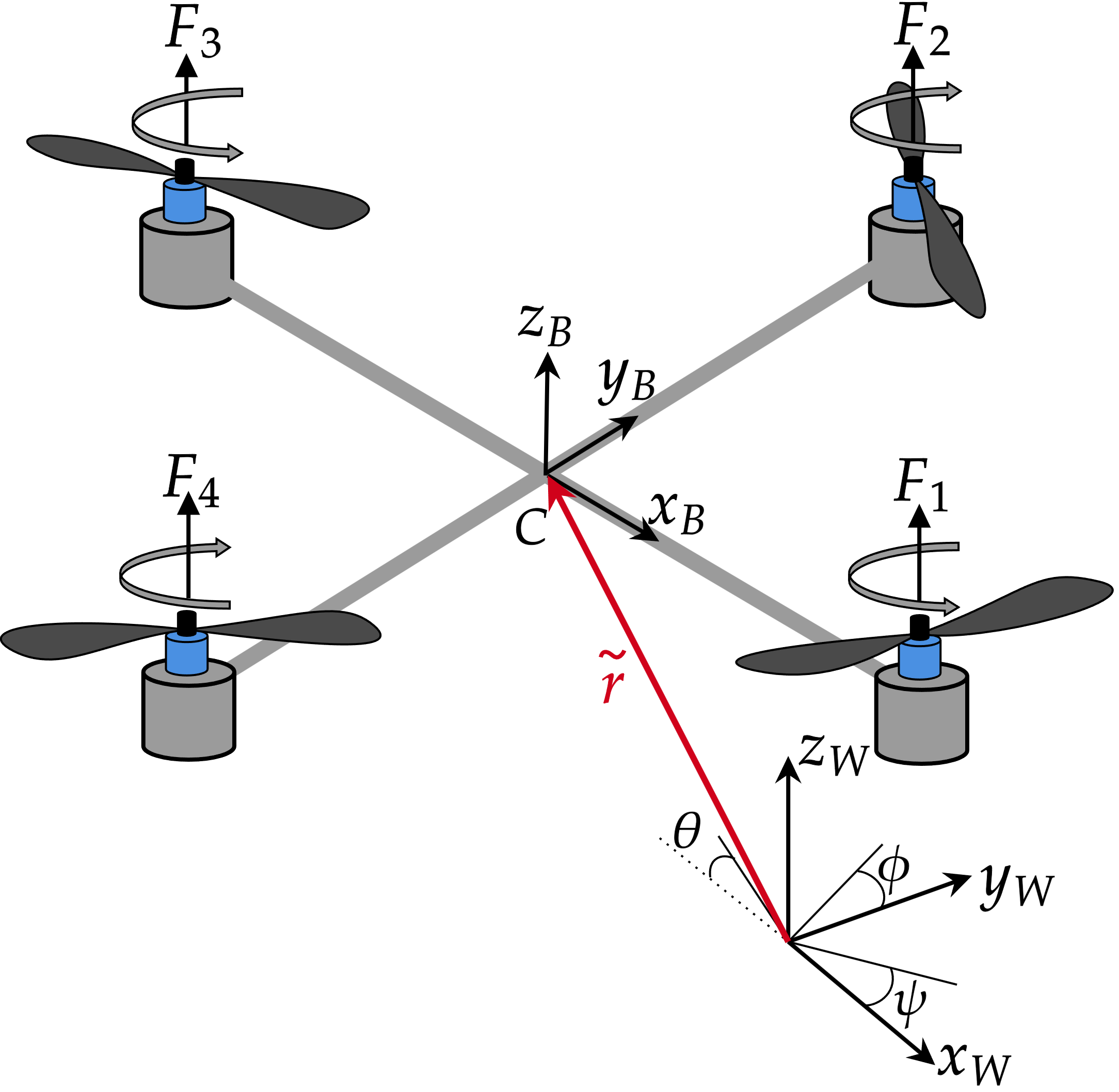}}
		\caption{Free body diagram and coordinate frames of the quadrotor.}
		\label{model}
	\end{figure}

	This section discusses the six degrees of freedom quadrotor model and the control architecture used for validation. Figure \ref{model} shows the free-body diagram of the quadrotor, where the coordinate systems $W$, and $B$ represent the world frame and body frame of the vehicle, respectively. Frame $B$ is attached to the vehicle's center of mass $C$. The world frame axes are $\{x_w, y_w, z_w\}$, and the body frame axes are $\{x_b, y_b, z_b\}$. The vector $\tilde{r}=[
	x(t) \quad y(t) \quad z(t)
	]^T\in \mathbb{R}^3$ is the position vector of $C$ in $W$. The attitude vector $[
	\phi(t)\quad\theta(t)\quad\psi(t)
	]^T\in \mathbb{R}^3$ indicates the angular position in the world frame. The pitch ($\theta$), roll ($\phi$), and yaw($\psi$) define the rotation about $y_w$, $x_w$, and $z_w$ axis, respectively.
The translational and rotational dynamic equations of the quadrotor \cite{Beard} are given by
\begin{equation}
	\begin{bmatrix}
		\Ddot{x}\\\Ddot{y}\\\Ddot{z}\end{bmatrix}=\frac{1}{m}\left(\begin{bmatrix} 	u_1( C_{\phi}S_{\theta}C_{\psi}+S_{\phi}S_{\psi})\\u_1(C_{\phi}S_{\theta}S_{\psi}-S_{\phi}C_{\psi})\\u_1C_{\phi}C_{\theta}
	\end{bmatrix}-\frac{1}{m}\begin{bmatrix} 0\\0\\mg\end{bmatrix}\right)
\end{equation}
\begin{equation}\label{6dof1}
	\begin{bmatrix}
		\Ddot{\phi}\\ \Ddot{\theta}\\ \Ddot{\psi}
	\end{bmatrix}=\begin{bmatrix} \frac{1}{J_{xx}}u_2\\ \frac{1}{J_{yy}}u_3 \\ \frac{1}{J_{zz}}u_4 \end{bmatrix}
\end{equation}	
where $m$, $g$, and $[
J_{xx} \quad J_{yy} \quad J_{zz}
]^T$ represent mass, standard gravitational acceleration, and inertia matrix, respectively. Here, $C_\phi$ and $S_\phi$ stand for $\cos\phi$ and $\sin\phi$, respectively, with similar notations applied to $\psi$ and $\theta$. The control inputs $u_1=\sum_{i=1}^{4}F_i$ is the total upward thrust acting along $z_B$-axis. The other control inputs $u_2$, $u_3$, and $u_4$ in Eq. \eqref{6dof1}  are the rolling moment, pitching moment, and yawing moment, respectively. 
The derivative of attitude angles related to angular velocity components in the body frame ($p,q,r$) as 
\begin{equation} \label{pqr}
	\begin{bmatrix}
		p\\q\\r
	\end{bmatrix}
	=\begin{bmatrix}
		1 & 0 & -\sin{\theta}\\0 & \cos{\phi} &\sin{\phi} \cos{\theta}\\0 & -\sin{\phi} &\cos{\phi} \cos{\theta}
	\end{bmatrix}
	\begin{bmatrix}
		\dot{\phi}\\ \dot{\theta}\\ \dot{\psi}
	\end{bmatrix}
\end{equation}
%\section{Trajectory Generation and Control}\label{TGC}
%This section describes the trajectory generation and the control framework of the quadrotor. 
The desired velocity components ($v_{xdes},v_{ydes},v_{zdes}$) are generated using the proposed guidance commands in 
Eqs. \eqref{pointmass}-\eqref{pointmassend} as  
\begin{align}\label{trajectory}
	v_{xdes} &=V \cos{\gamma_{des}(t)}\cos{\chi_{des}(t)}\\
	v_{ydes} &=V \cos{\gamma_{des}(t)} \sin\chi_{des}(t)\\
	v_{zdes} &=V \sin{\gamma_{des}(t)}\label{traj}
\end{align}
Using Eqs. \eqref{trajectory}-\eqref{traj} the desired position $(x_{des}, y_{des},z_{des})$ can be generated as
\begin{align}\label{xdes}
	x_{des}&=x(t_i)+\int_{t_i}^{t}{v}_{xdes} dt\\ y_{des}&=y(t_i)+\int_{t_o}^{t}{v}_{ydes} dt\\
	z_{des}&=z(t_i)+\int_{t_o}^{t}{v}_{zdes} dt	\label{zdes}	
\end{align}
 \begin{figure}[htbp]
	\centerline{\includegraphics[width=87mm]{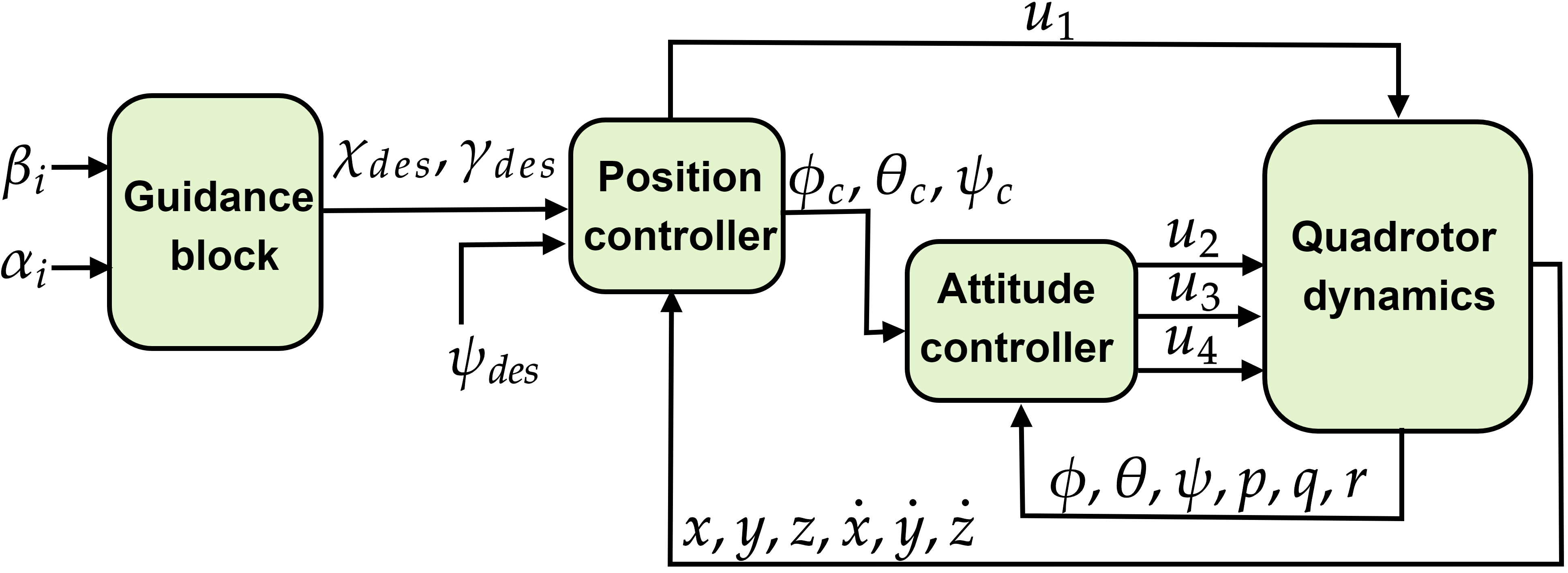}}
	\caption{Guidance and control architecture}
	\label{3D:blockdiagram}
\end{figure}
\begin{table*}[h!]
	\centering
	\caption{Qualitative comparison with existing methods}\label{tableofcomparison}	
	\begin{tabular}{llll}
		\hline
		\multicolumn{1}{c}{\begin{tabular}[c]{@{}c@{}}Traversal \\ methods\end{tabular}} & \multicolumn{1}{c}{\begin{tabular}[c]{@{}c@{}}Window information \\ required\end{tabular}}                                                                                   & \multicolumn{1}{c}{Traversal trajectory planning computations}                                                                                                                                                                                                                                                                                    & \multicolumn{1}{c}{\begin{tabular}[c]{@{}c@{}}Flexibility in\\ replanning the\\ traversal path\end{tabular}} \\ \hline
		\begin{tabular}[c]{@{}l@{}}Mellinger et al. \\ \cite{b3}\end{tabular}           & \begin{tabular}[c]{@{}l@{}}Prior information about \\ window dimension in \\ world frame\end{tabular}                                                                        & \begin{tabular}[c]{@{}l@{}}3-D trajectory and  corresponding velocity  components are computed\\ through backward integration of the dynamic equations of quadrotor\\ motion using a known final goal location and specified traversal\\ time duration.\end{tabular}                                                                          & No                                                                                                           \\ \hline
		\begin{tabular}[c]{@{}l@{}}Mellinger and \\ Kumar \cite{b4}\end{tabular}     & \begin{tabular}[c]{@{}l@{}}Prior information about \\ window dimension in \\ world frame\end{tabular}                                                                        & \begin{tabular}[c]{@{}l@{}}Computation of time-parameterized optimal traversal trajectory\\ through a series of known waypoints  with the window\\ clearance constraints. The coefficients of these piecewise time-\\parameterized polynomial trajectories are evaluated by optimizing \\the integral of the squared norm of the snap.\end{tabular} & No                                                                                                           \\ \hline
		\begin{tabular}[c]{@{}l@{}}Upadhay et al. \\ \cite{upadhyay2022generation}\end{tabular}            & \begin{tabular}[c]{@{}l@{}}Prior information about \\ window dimension in \\ world frame\end{tabular}                                                                        & \begin{tabular}[c]{@{}l@{}}3-D traversal trajectory is computed through geometric interpolation\\ using four-parameter logistic polynomials and coefficients of these \\ time-parameterized polynomials are computed by considering\\ window clearance constraints.\end{tabular}                                                              & No                                                                                                           \\ \hline
		\begin{tabular}[c]{@{}l@{}}Loiano et al. \\ \cite{estimation}\end{tabular}             & \begin{tabular}[c]{@{}l@{}}Window pose information\\ as computed through\\ image features\end{tabular}                                                                     & \begin{tabular}[c]{@{}l@{}}Trajectory generation is similar to the one in \cite{b4}.\end{tabular}                                                                                                                                                                                                                                                & No                                                                                                           \\ \hline
		\begin{tabular}[c]{@{}l@{}}Wang et al. \\ \cite{MPCC}\end{tabular}              & \begin{tabular}[c]{@{}l@{}}Prior information about \\ window dimension in \\ world frame\end{tabular}                                                                                      & \begin{tabular}[c]{@{}l@{}}2-D trajectory is computed using sigmoid function-based geometric\\ interpolation. The solution is valid for limited planar scenarios.\end{tabular}                                                                                                                                                                & Yes                                                                                                          \\ \hline
		\begin{tabular}[c]{@{}l@{}}Falanga et al. \\ \cite{activevision}\end{tabular}            & \begin{tabular}[c]{@{}l@{}}Centroid of the window as \\ located using eight corner \\ points in the image,\\ window depth information\end{tabular}                & \begin{tabular}[c]{@{}l@{}}Computation of time-parameterized second-order trajectory and\\ velocity profiles passing through the window centroid.\\ Trajectory parameters are computed by solving minimum traversal\\ time optimization problem.\end{tabular}                                                                      & Yes                                                                                                          \\ \hline
		\begin{tabular}[c]{@{}l@{}}Guo and \\ Leand \cite{Guo}\end{tabular}          & \begin{tabular}[c]{@{}l@{}}Pre-captured images of the\\ window need to be stored \\ in advance to facilitate\\ comparison with real-time \\images, window depth \\information\end{tabular} & \begin{tabular}[c]{@{}l@{}}Image-based trajectory planning is utilized wherein the trajectory is\\ expressed in terms of invariant visual features and vehicle's heading \\ direction. A dynamically feasible trajectory is generated by solving  \\ quadratic programming optimization problem, which minimize the \\ snap of relative translation between the current and desired images.\end{tabular}                                                                              & Yes                                                                                                          \\ \hline
		\begin{tabular}[c]{@{}l@{}}Mueller et al. \\ \cite{QuadrotorGuidanceimage}\end{tabular}            & \begin{tabular}[c]{@{}l@{}}Original image of the window\\is needed to conduct image\\ matching algorithm on real-\\time images\end{tabular}                                            & \begin{tabular}[c]{@{}l@{}}Trajectory generation using desired thrust and angular velocities for\\ lateral traversal through the window. A model predictive control \\approach is used to numerically obtain the solution. \end{tabular}                                                                                                                                                                                             & Yes                                                                                                          \\ \hline
		\begin{tabular}[c]{@{}l@{}}Midhun and \\ Ratnoo  \cite{mekjrnl}\end{tabular}      & \begin{tabular}[c]{@{}l@{}}Bearing information of gap \\ extremities\end{tabular}                                                                                            & \begin{tabular}[c]{@{}l@{}}Trajectory generation using desired heading guidance command. \\The solution is limited to planar scenarios.\end{tabular}                                                                                                                                                                              & Yes                                                                                                          \\ \hline
		\begin{tabular}[c]{@{}l@{}}Proposed \\ method\end{tabular}                       & \begin{tabular}[c]{@{}l@{}}Bearing information of the \\ window vertices\end{tabular}                                                                                        & \begin{tabular}[c]{@{}l@{}}The guidance inputs are characterized by commanded flight path and \\heading angles in Eqs. \eqref{desg} and \eqref{desheading}, respectively. \end{tabular}                                                                                                                                                                                                                  & Yes                                                                                                          \\ \hline
	\end{tabular}
\end{table*}
where $[x(t_i)\quad y(t_i)\quad z(t_i)]^T\in\mathbb{R}^3$ represents vehicle’s initial position in the world frame $W$. 
The guidance and control architecture of the quadrotor is depicted in Fig. \ref{3D:blockdiagram}, where the desired flight path and heading angles are generated through the guidance block utilizing bearing angle information.
As given in Ref. \cite{mekjrnl}, the desired position Eqs. \eqref{xdes}-\eqref{zdes} and its derivatives Eqs. \eqref{trajectory}-\eqref{traj} can be related to the control inputs $u_1$, $u_2$, $u_3$, and $u_4$ through control relations Eqs. \eqref{u1}-\eqref{u4}. 
\begin{align}
	u_1 &=m(g+K_{dz}(v_{zdes}-\dot{z})+K_{pz}(z_{des}-z))\label{u1}\\
	u_2&=K_{d\phi}(p_c-p)+K_{p\phi}(\phi_c-\phi)\\ \label{u2}
	u_3&=K_{d\theta}(q_c-q)+K_{p\theta}(\theta_c-\theta)\\ 
	u_4&=K_{d\psi}(r_c-r)+K_{p\psi}(\psi_c-\psi)  \label{u4}	
\end{align}
Here, $(K_{dz}, K_{pz})$, $(K_{d\phi}, K_{p\phi})$, $(K_{d\theta}, K_{p\theta})$, and  $(K_{d\psi}, K_{p\psi})$  represent the tunable controller gains for $u_1$, $u_2$, $u_3$, and $u_4$, respectively, and the commanded angular velocity components ($p_c,q_c,r_c$) are computed using the derivatives of commanded attitude angles ($\dot{\phi}_c,\dot{\theta}_c,\dot{\psi}_c$) using Eq. \eqref{pqr}.
\section{QUALITATIVE COMPARISON WITH EXISTING TRAVERSAL METHODS}\label{3D:COMP}
This section presents a qualitative comparison study of the proposed method with existing window traversal methods. The comparison is structured around three key criteria: (1) Nature of window information required to generate guidance commands; 
(2) Computational complexity; and
(3) Flexibility in re-planning the traversal path.
Table \ref{tableofcomparison} summarizes the comparative remarks. It can be noted that, the proposed traversal method offers significant advantages due to its reliance only on relative bearing information of window vertices and simpler trajectory planning computations, making it a promising approach for real-time applications. In addition, the proposed traversal trajectories are dynamically generated based on instantaneous bearing
angles, enabling the flexibility of incorporating updated information as the flight progresses.
\section{NUMERICAL SIMULATION RESULTS}\label{Numerical}
Numerical simulations are carried out to evaluate the performance of the proposed guidance method using a six degrees of freedom quadrotor vehicle model with mass and moment of inertia properties as $m=0.47$ kg, $J_{xx}=0.0086$ kg-m$^2$, $J_{yy}=0.0086$ kg-m$^2$, and $J_{zz}= 0.0176$ kg-m$^2$. A window with dimensions $a=4$ m and $b=3$ m with vertices $E_1=(12,15,13)$ m, $E_2=(16,15,13)$ m, $E_3=(16,15,10)$ m, and $E_4=(12,15,10)$ m, is considered. The centroid of the window is $W_c=(14,15,11.5)$ m, which is the desired traversal point. 
The quadrotor's traversal speed is set to $V = 1$ m/s for all simulations, with an initial velocity of 0 m/s and the desired yaw angle is $\psi_{des} = 0$ deg.
The controller gains for position and attitude control are selected as $K_{pz}=3.8$, $K_{dz}=3.5$, $K_{py}=12.7$, $K_{dy}=4.2$, $K_{px}=6$, $K_{dx}=3.5$, $K_{pjx}=6$, $K_{djx}=3.5$, $K_{pjy}=12.7$, $K_{djy}=4.2$, $K_{p\phi}=12.8$, $K_{d\phi}=0.5$, $K_{p\theta}=1.8$, $K_{d\theta}=0.2$, $K_{p\psi}=2$, and $K_{d\psi}=0.5$. 
Once the traversal condition in Eq. \eqref{c2} is met, the guidance commands follow their already achieved heading angle $\chi_{des}=\frac{\pi}{2}$ rad and flight path angle $\gamma_{des}=0$ rad. The commanded pitch and roll angles of the quadrotor are saturated to be within $\pm20$ deg.
\subsection{Case 1: Sample scenario}
\begin{figure}[htbp]
	\centerline{\includegraphics[width=88mm,trim={0.75cm 1.45cm 0.75cm 2.00cm},clip]{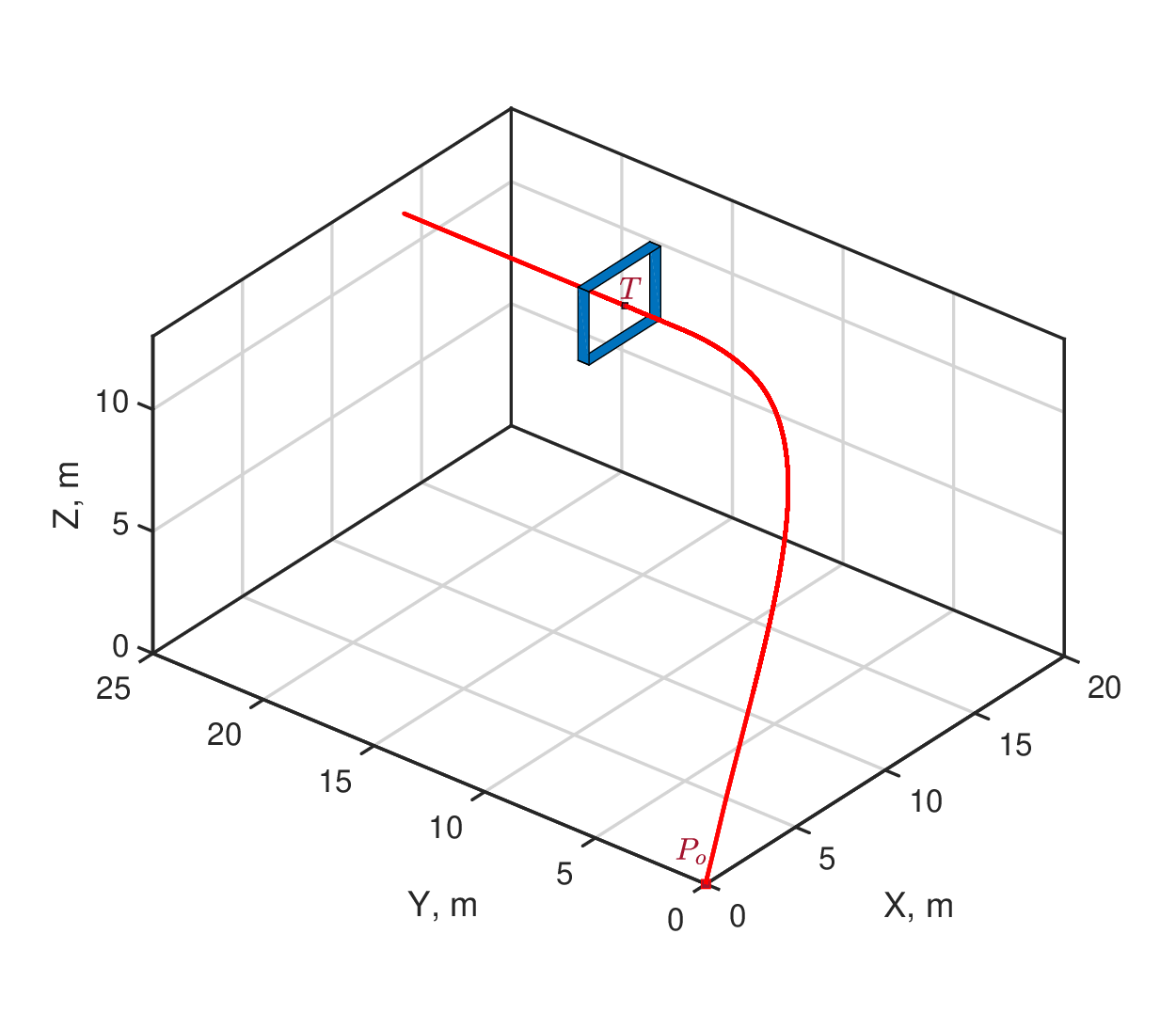}}
	\caption{Quadrotor trajectory for Case 1.}
	\label{case1trajectory}
\end{figure}
\begin{figure}[htbp] 
	\centering
	\subfloat[{} \label{case1bisector1}]
	{\includegraphics[width=8cm,trim={0cm 0.08cm 0.5cm 0.4cm},clip]{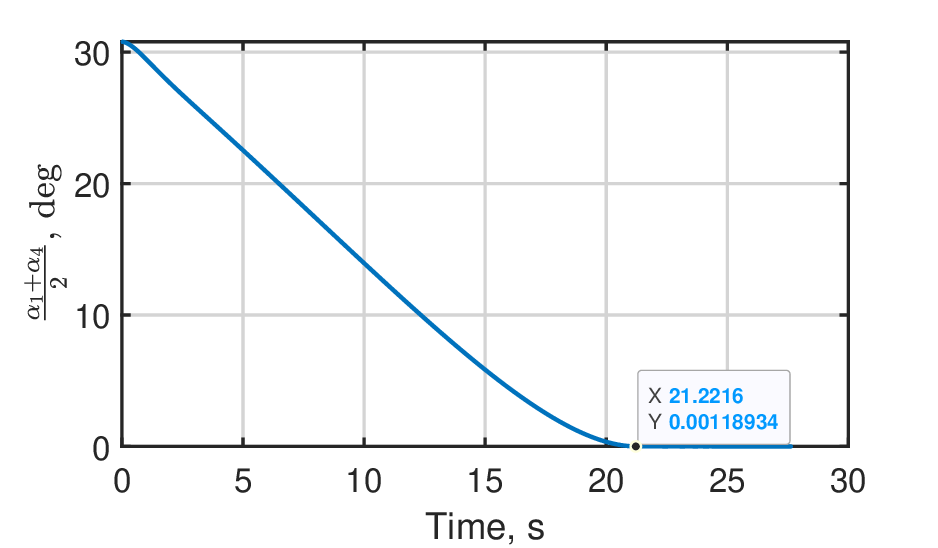}}\\
	\subfloat[{}  \label{case1bisector2}]
	{\includegraphics[width=8cm,trim={0cm 0.08cm 0.5cm 0.4cm},clip]{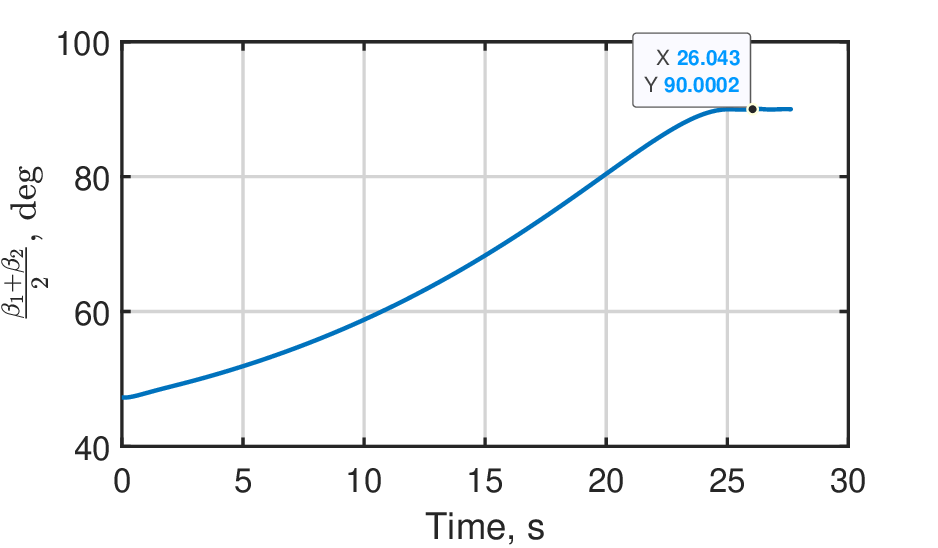}}\\
	\subfloat[{} \label{sgamma}]
	{\includegraphics[width=8cm,trim={0cm 0.03cm 0.5cm 0.4cm},clip]{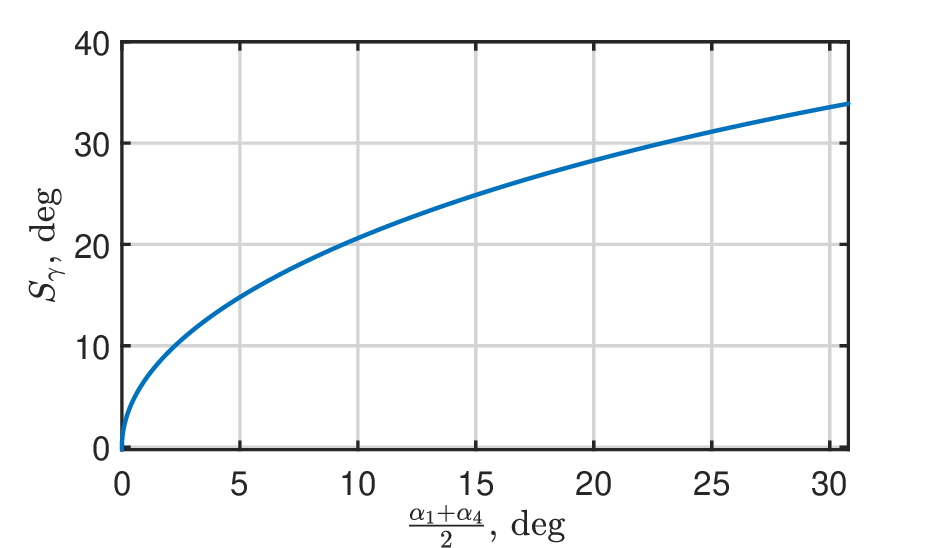}}\\
	\subfloat[{} \label{schi}]
	{\includegraphics[width=8cm,trim={0cm 0.03cm 0.5cm 0.4cm},clip]{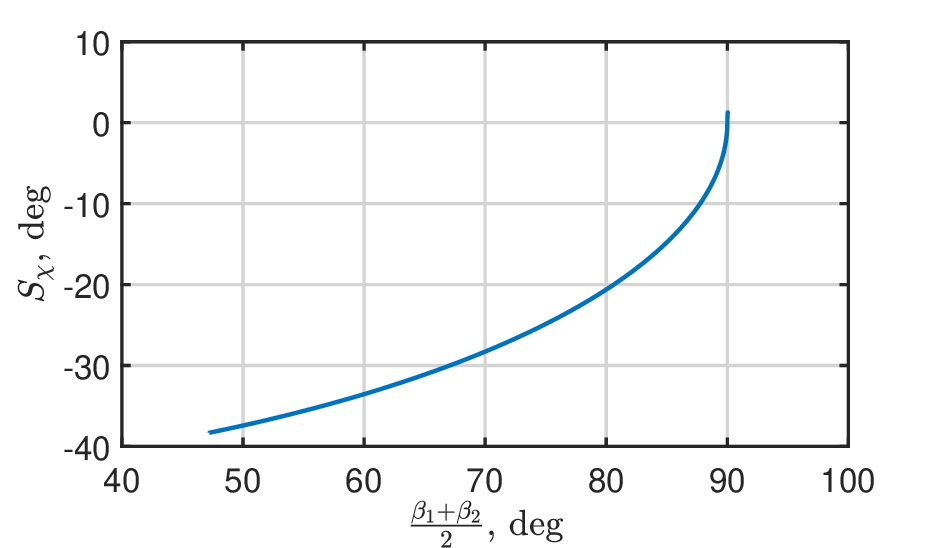}}\\
	\caption{Results for Case 1: (a) Angular bisector of elevation angles. (b) Angular bisector of azimuth angles. (c) Shaping angle component for flight path angle. (d) Shaping angle component for heading angle.}
\end{figure}
This case considers guided trajectory generation using noise-free bearing angles: $\alpha_1$, $\alpha_4$, $\beta_1$, and $\beta_2$.
 Quadrotor's initial location $P_o$ is set to be at origin. The resulting quadrotor trajectory depicted in Fig. \ref{case1trajectory} shows that the quadrotor successfully traverses through the desired traversal point $T(14,15,11.5)$ m at $27.62$ s. 
The angular bisector of elevation angles $\frac{\alpha_1+\alpha_4}{2}$ converges its desired final value of zero degrees, as depicted in Fig. \ref{case1bisector1}. Likewise, the angular bisector of azimuth angles $\frac{\beta_1+\beta_2}{2}$ achieves the desired value of $\frac{\pi}{2}$ rad before reaching the traversal point, as shown in Fig. \ref{case1bisector2}. Figure \ref{sgamma} shows the variation of the shaping angle component $S_\gamma$ plotted against the angular bisector $\frac{\alpha_1+\alpha_4}{2}$. In accordance with Eq. \eqref{des2}, $S_\gamma$ decreases monotonically to zero as the angular bisector of elevation angles, $\frac{\alpha_1+\alpha_4}{2}$ reaches zero, facilitating the desired traversal. Further, Fig. \ref{schi} shows the variation of the shaping angle $S_\chi$ with the angular bisector component $\frac{\beta_1+\beta_2}{2}$. As desired, the shaping component reaches zero when the angular bisector $\frac{\beta_1+\beta_2}{2}$ attains its final desired value of $\frac{\pi}{2}$. 
\begin{figure}[] 
	\centering
	\subfloat[{} \label{roll}]
	{\includegraphics[width=8cm,trim={0cm 0.08cm 0.5cm 0.4cm},clip]{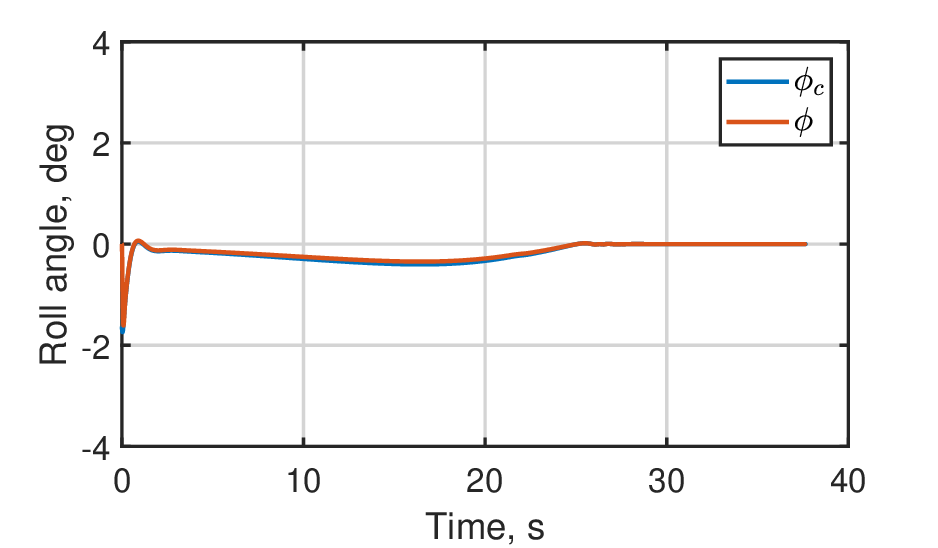}}\\
	\subfloat[{}  \label{pitch}]
	{\includegraphics[width=8cm,trim={0cm 0.08cm 0.5cm 0.4cm},clip]{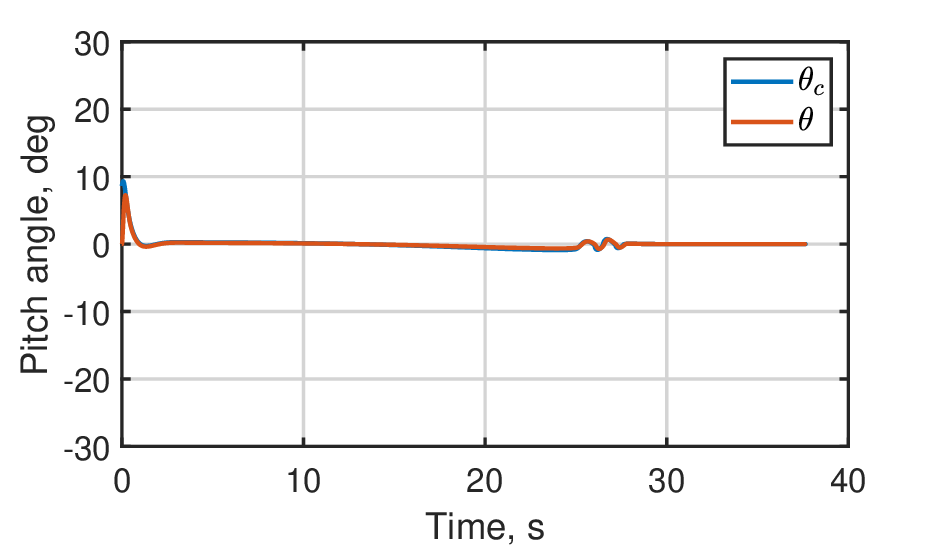}}\\
	\subfloat[{}  \label{velocity}]
	{\includegraphics[width=8cm,trim={0cm 0.08cm 0.5cm 0cm},clip]{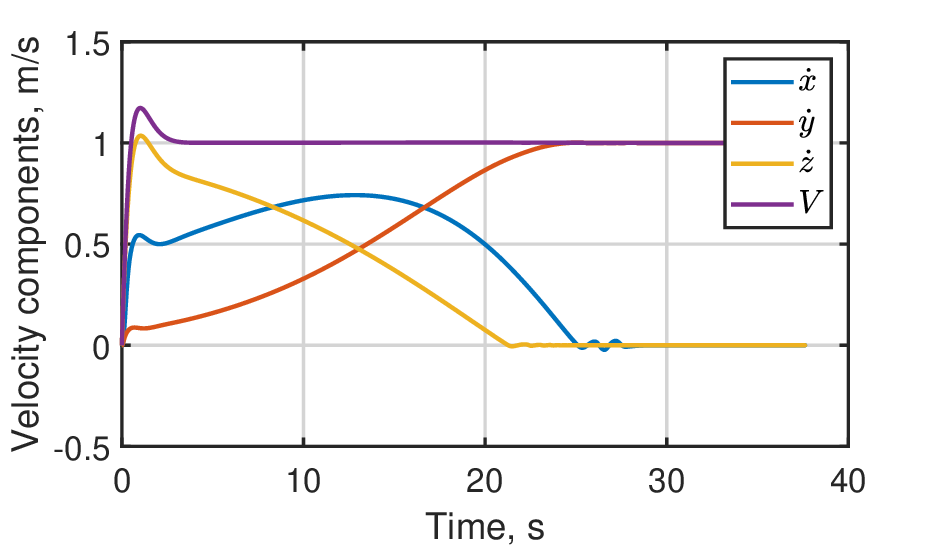}}
	\caption{Results for Case 1: (a) Roll angle response. (b) Pitch angle response. (c) Quadrotor velocity components.} 
	\label{case1attitude}
\end{figure}
The roll and pitch responses of the quadrotor are illustrated in Figs. \ref{roll} and \ref{pitch}, respectively. The velocity components of the quadrotor are plotted in Fig. \ref{velocity}.
\subsection{Case 2: Presence of noise in bearing measurements}
This case study investigates the performance of the proposed guidance method when the bearing angle information is corrupted by noise.
The noisy bearing measurements are generated by adding zero-mean Gaussian noise to the true values. That is, $\alpha_{im}(t)=\alpha_i(t)+\mathcal{N}(0,\sigma^{2})$, and $\beta_{im}(t)=\beta_i(t)+\mathcal{N}(0,\sigma^{2})$, where $\alpha_{im}$ and $\beta_{im}$ represent the noisy bearing measurements.
\begin{figure}[htbp] 
	\centering
	\subfloat[{} \label{case2trajectory}]
	{\includegraphics[width=85mm,trim={0.5cm 1cm 0.5cm 1.8cm},clip]{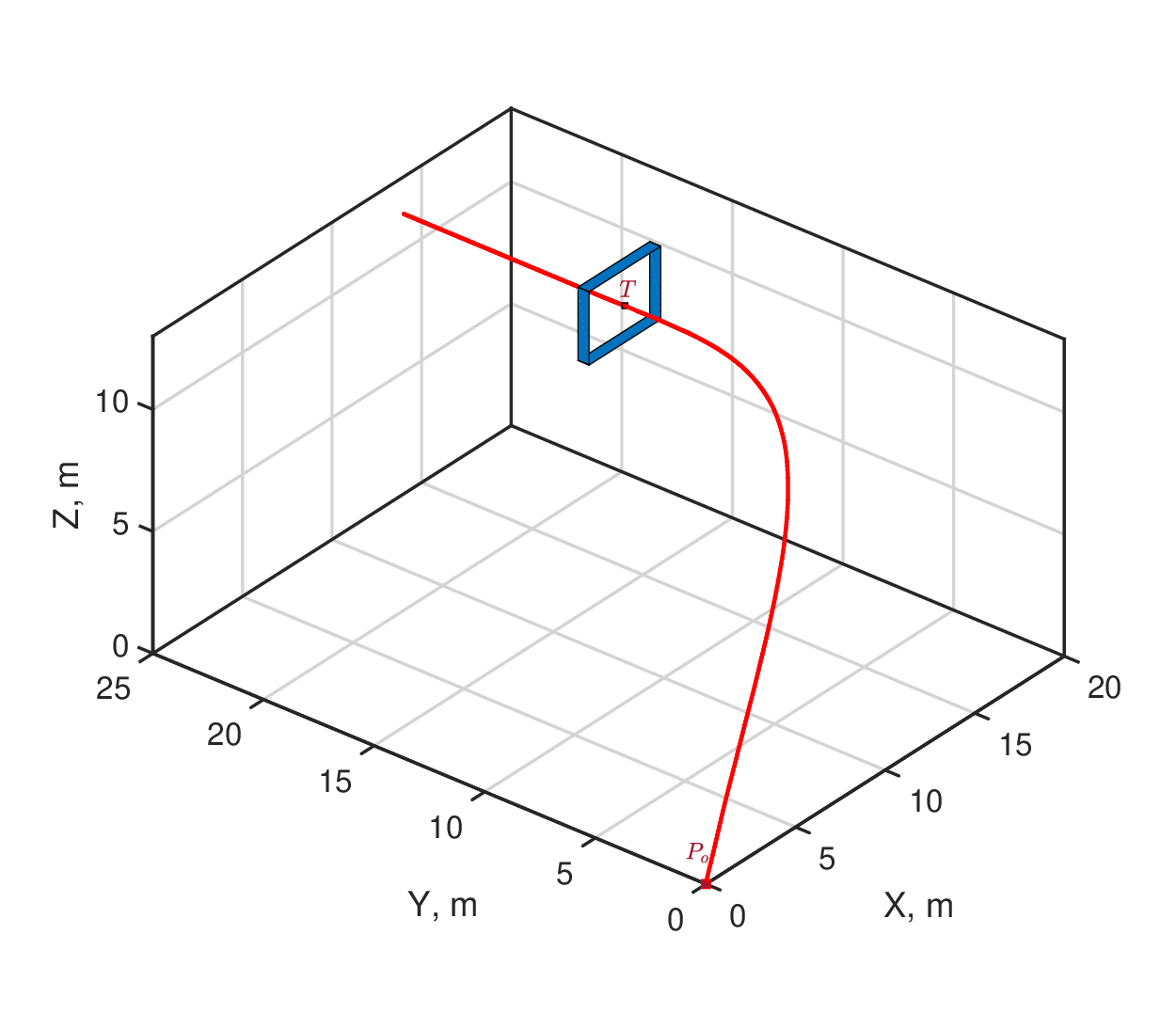}}\\
	\subfloat[{} \label{elev}]
	{\includegraphics[width=8cm,trim={0.2cm 0.08cm 0.5cm 0.4cm},clip]{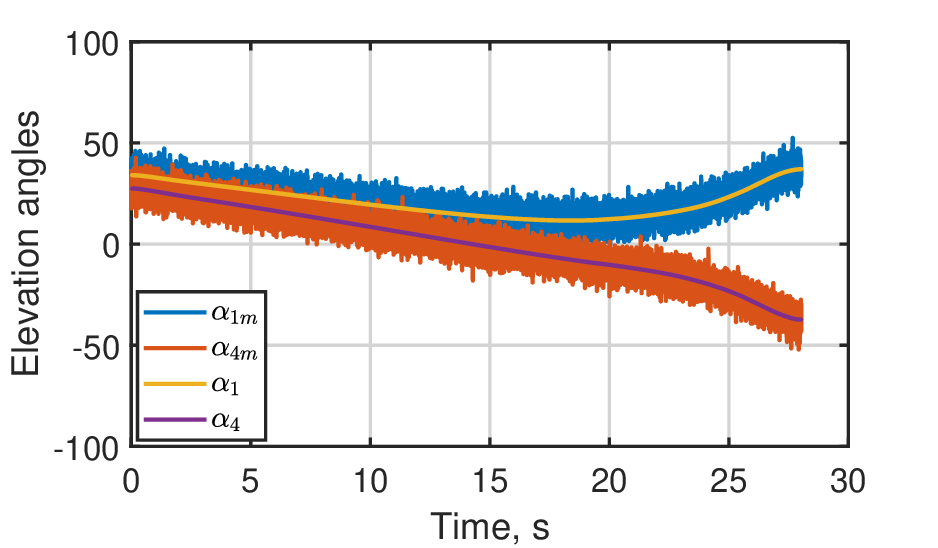}}\\
	\subfloat[{}  \label{azim}]
	{\includegraphics[width=8cm,trim={0.2cm 0.08cm 0.5cm 0.4cm},clip]{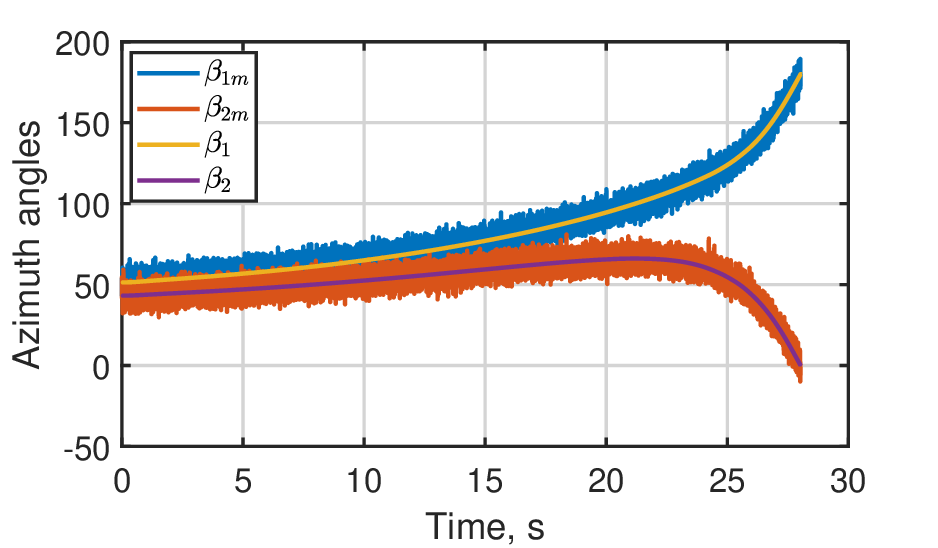}}\\
	\subfloat[{}  \label{dist}]
	{\includegraphics[width=8cm,trim={0.2cm 0.08cm 0.5cm 0cm},clip]{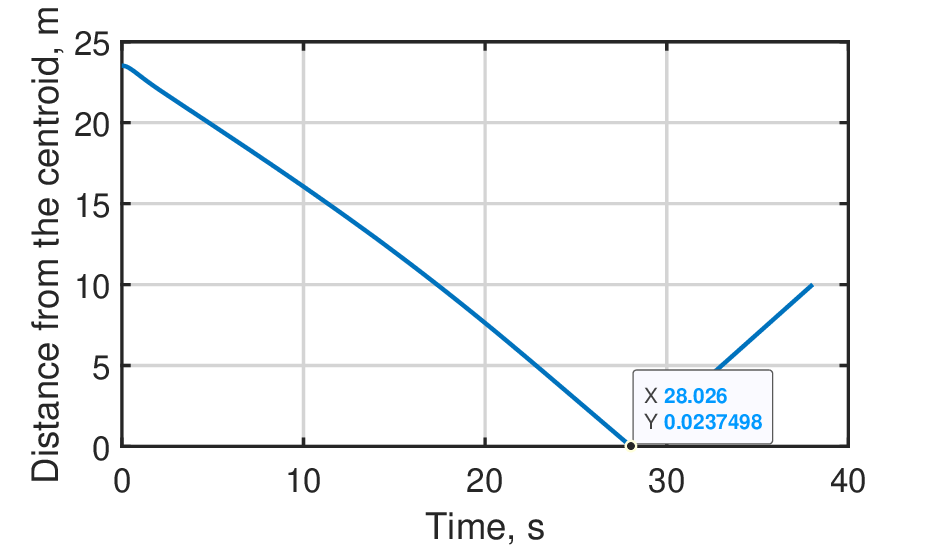}}
	\caption{Results for Case 2: (a) Quadrotor trajectory. (b) Elevation angles. (c) Azimuth angles. (d) Distance from the centroid.}
	\label{case2bearing}
\end{figure}
 Here, $\sigma$ is the standard deviation associated with the noise. The robustness of the proposed traversal method is studied with respect to unfiltered $\alpha_{im}$ and $\beta_{im}$ utilized in the proposed guidance commands of Eqs. \eqref{desg} and \eqref{desheading}. The initial conditions and window geometry are identical to the ones considered in Case 1. 
Further, this study considers the standard deviation of the noise to be $\sigma=4$ deg. The resulting quadrotor trajectory plotted in Fig. \ref{case2trajectory}, indicates safe traversal of the quadrotor through the window.
True values and measured values of the elevation angles and azimuth angles are shown in Figs. \ref{elev} and \ref{azim}, respectively. The vehicle distance from the centroid of the window is shown in Fig. \ref{dist}, which indicates that the quadrotor reaches the traversal point with a negligible error of $0.023$ m from its desired value.  
\subsection{Case 3: Various initial conditions and noisy bearing measurements}
A Monte Carlo simulation study is performed with hundred different initial positions of the quadrotor satisfying $x(t_i)\in[0,30]$ m, $y(t_i)\in[0,14]$ m, and $z(t_i)\in[0,20]$ m.
\begin{table}[htbp]
	\centering
	\caption{Statistical results of Case 3}\label{3D:tab1}
	\begin{tabular}{ccc}
		\hline
		\hline
		\multirow{2}{*}{\begin{tabular}[c]{@{}c@{}}Bearing measurement noise\\  $\mathcal{N}$(0, $\sigma^2$)\end{tabular}} & \multicolumn{2}{c}{\begin{tabular}[c]{@{}c@{}}Distance between traversal point \\and centroid (m)\end{tabular}} \\ \cline{2-3} 
		& Mean                                    & Standard deviation                                 \\ \hline
		$\mathcal{N}(0,1^{2})$                                                                                                                                   & 0.0351                                 & 0.0694                                             \\ \hline
		$\mathcal{N}(0,2^{2})$                                                                                                                                   & 0.0488                                  & 0.0859                                             \\ \hline
		$\mathcal{N}(0,3^{2})$                                                                                                                                   & 0.0568                                  & 0.0945                                             \\ \hline
		$\mathcal{N}(0,4^{2})$                                                                                                                                   & 0.0571                                  & 0.0745                                            \\ \hline
		$\mathcal{N}(0,5^{2})$                                                                                                                                    & 0.0447                                  & 0.0496                                             \\ \hline
		$\mathcal{N}(0,6^{2})$                                                                                                                                    & 0.0618                                  & 0.0805                                             \\ \hline
		$\mathcal{N}(0,7^{2})$                                                                                                                                    & 0.0654                                  & 0.1076                                           \\ \hline
		\hline
	\end{tabular}
\end{table}
\begin{figure}[htbp]
	\centerline{\includegraphics[width=90mm,,trim={0.5cm 0.5cm 0.5cm 1cm},clip]{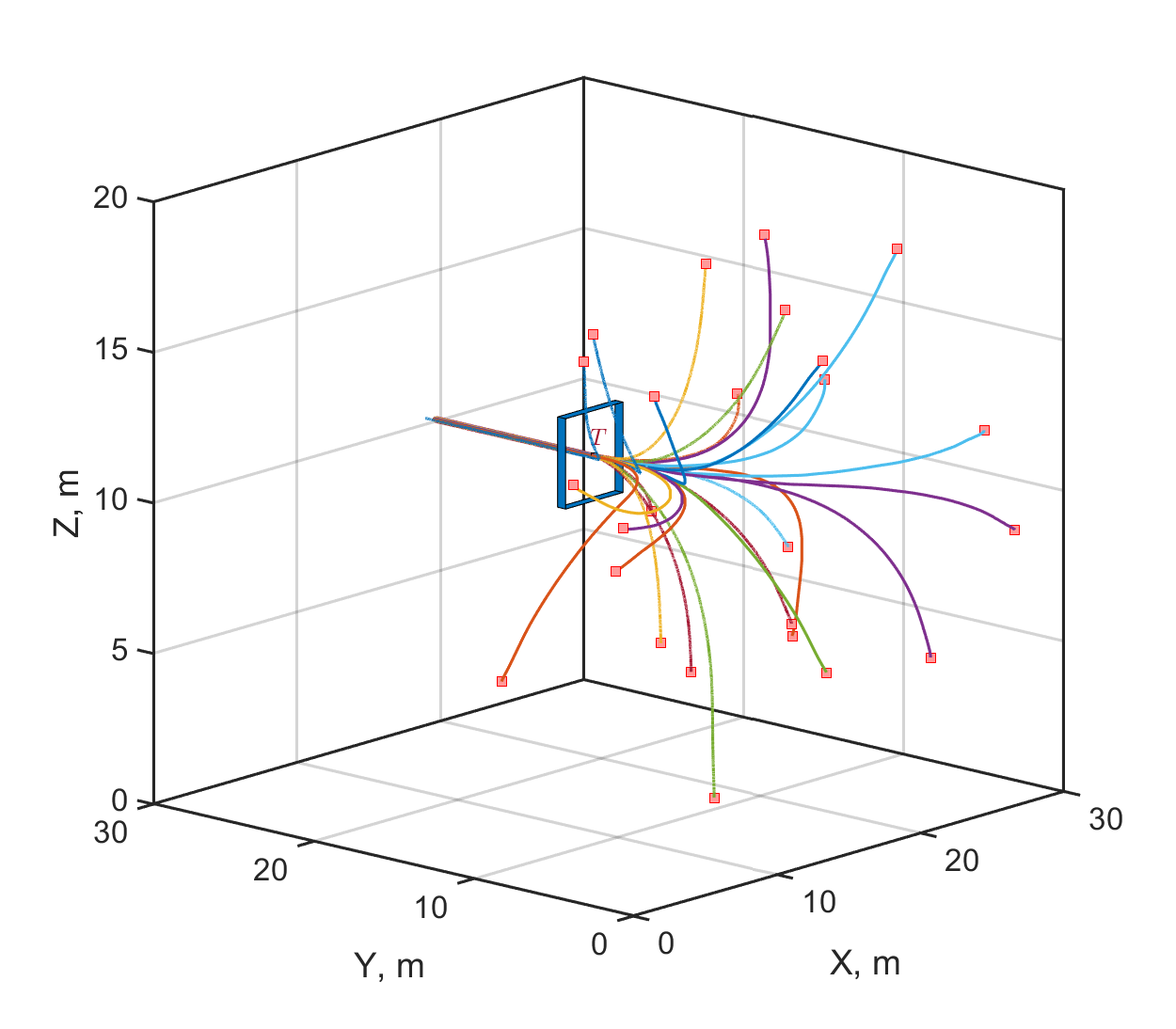}}
	\caption{Twenty quadrotor trajectories from the study considered in Case 3.}
	\label{mic}
\end{figure} 
The bearing measurements are corrupted by noise with its characteristics similar to those in Case 2 with $\sigma=1,2,\ldots, 7$ deg, respectively. The performance of the proposed method is assessed by examining the error between the desired and achieved traversal point. The results are illustrated in Table. \ref{3D:tab1}.
The mean and standard deviation results indicate a negligible deterioration of performance in passing through the centroid of the window with an increase in standard deviation in noise. 
Figure \ref{mic} plots 20 randomly selected trajectories from this study.  
\section{Experimental Results} \label{Exprmnt}
\begin{figure*}[t]
	\centering
	\includegraphics[width=0.925\linewidth]{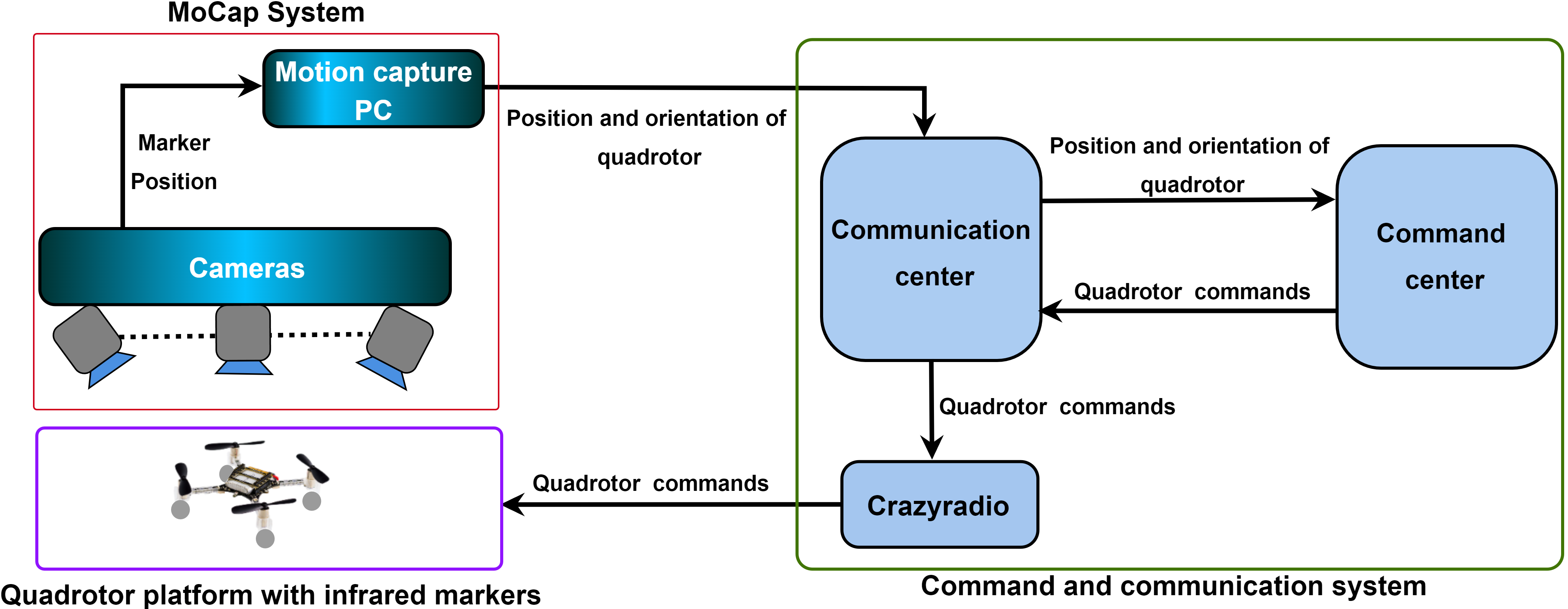}
	\caption{System architecture}
	\label{systemsrchitecture}
\end{figure*}
\begin{figure*}[htbp]
	\centering
	\includegraphics[width=14.5cm]{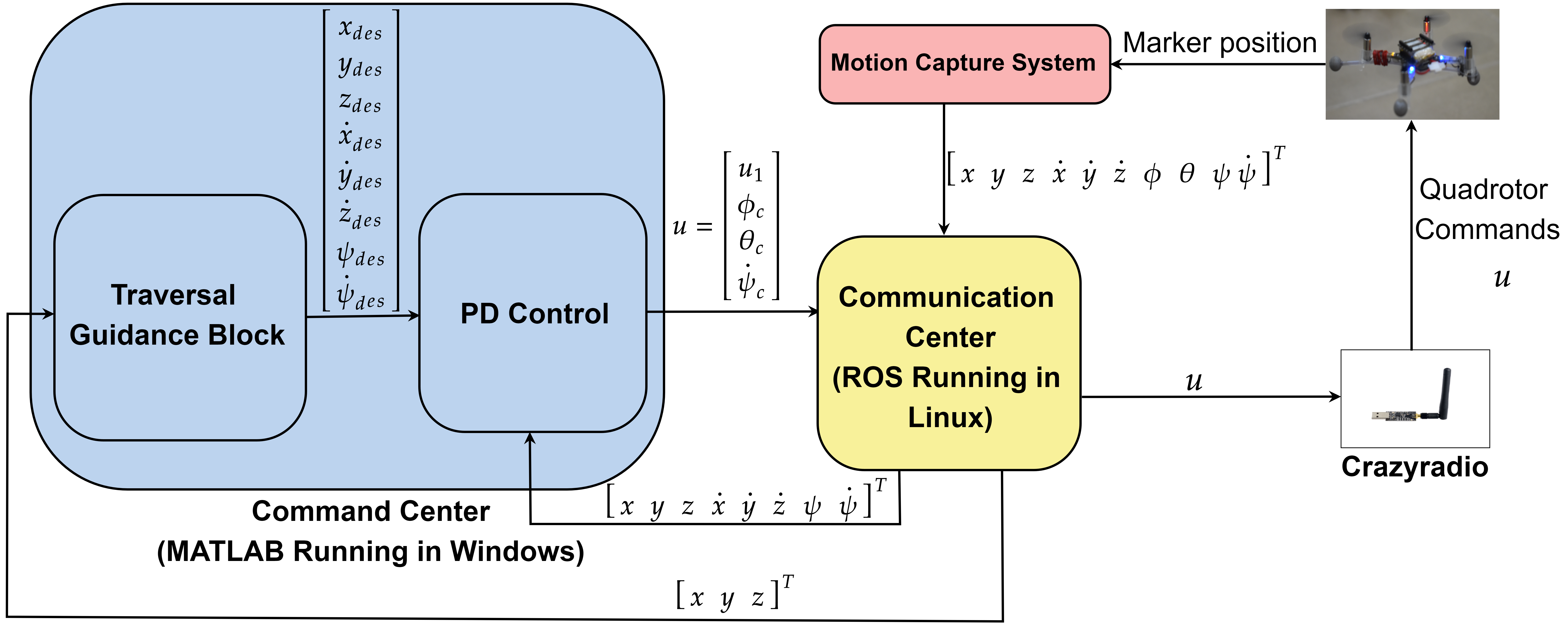}
	\caption{Details of command and communication system and its inter linkages}
	\label{Exp:Simulink}
\end{figure*}
\begin{figure}[htbp]
	\centering
	\includegraphics[width=6cm]{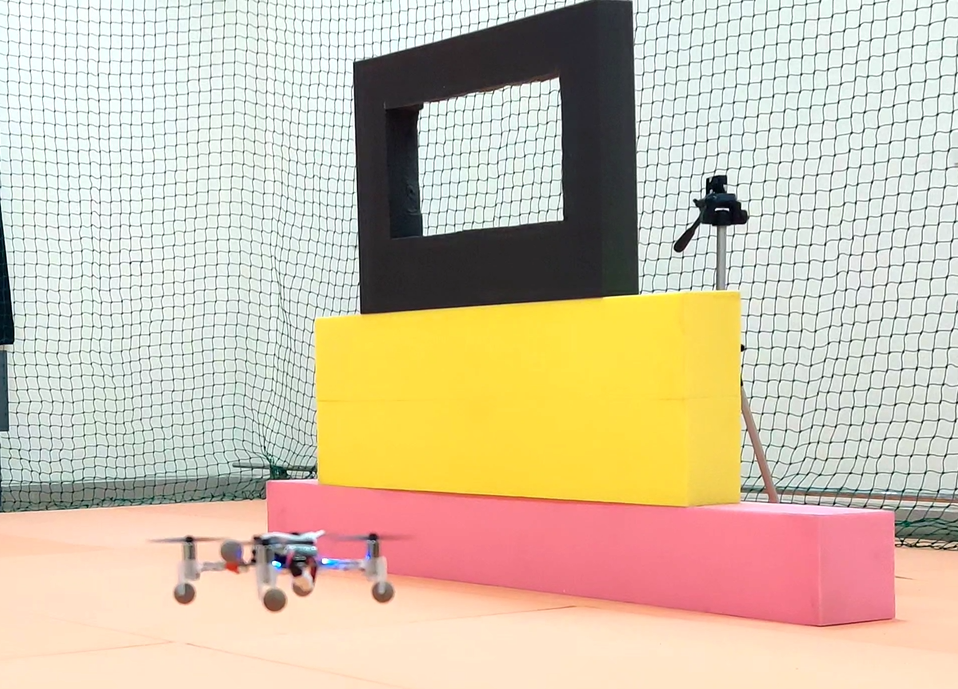}
	\caption{Crazyflie maneuvering towards the window}
	\label{window:snip}
\end{figure}
This section presents experimental validation results for the proposed guidance method. An indoor experimental setup is utilized, integrating a Crazyflie 2.0 nano quadrotor, motion capture system, and command and communication system.  Figure \ref{systemsrchitecture} provides an overview of the system architecture of the overall experimental setup, illustrating the nature of information exchange between the components.
 Crazyflie 2.0 nano quadrotor is an open-source experimental platform widely used in academic research. The dimension of the Crazyflie unit is 92 mm between diagonally opposed motor shafts and 29 mm in height. The total weight of the unit is 27 g. More information on Crazyflie 2.0 can be found in \cite{bitcraze2023}.

The position and orientation of the Crazyflie are measured using an indoor motion capture system, which utilizes sixteen OptiTrack Prime-13 cameras. Motion capture system uses infrared reflective markers attached to the quadrotor body. The LED illumination
ring around the camera produces infrared light, and the markers reflect it. The camera captures reflected infrared light and the motion capture software receives the data through an ethernet protocol connected to the cameras. The OptiTrack Motive software utilizes the data and compute the position of the markers placed on the quadrotor.
Consider an indicative window as shown in Fig. \ref{window:snip}.
In this setup, the bearing information of window vertices is obtained by placing reflective infrared markers on them.The results constitute a representative framework where the bearing information is obtained through external motion capture system and not from onboard sensors. Nevertheless, the flight trial results showcase the effectiveness of the proposed trajectory generation solution.

 The command and communication system comprises command center, communication center, and the Crazyradio communication dongle. Figure. \ref{Exp:Simulink} illustrates the details of these components and their inter linkages. 	
 The setup uses MATLAB Simulink as the command center, which runs on a Windows PC. As shown in Fig. \ref{Exp:Simulink},
	the proposed guidance methodology is implemented in the ``Traversal Guidance Block'' subsystem, which generates the desired quadrotor states $[x_{des}\;y_{des}\;z_{des}\; \dot{x}_{des}\;\dot{y}_{des}\;\dot{z}_{des} \; \psi_{des}\; \dot{\psi}_{des}]^T$. The quadrotor states $[x\;y\;z\; \dot{x}\;\dot{y}\;\dot{z} \; \psi\; \dot{\psi}]^T$ are obtained from the motion capture subsystem through the communication center. These feedback states are utilized by the ``PD Control'' subsystem, functioning as the autopilot for executing guidance commands while maintaining vehicle stability. 
	The Crazyflie quadrotor platform operates with four distinct control inputs, namely  total thrust, commanded roll, commanded pitch, and commanded yaw rate. Accordingly,
	\begin{equation}
		u=\begin{bmatrix}
			u_1 & \phi_c & \theta_c &\dot{\psi}_c 
		\end{bmatrix}^T
	\end{equation}
The communication center communicates $u$ to the quadrotor via Crazyradio dongle and it runs on a Linux PC, which uses Robotic Operating System (ROS). The communication center communicates with Crazyflie, motion capture system, and the command center in a fashion as shown in Figs. \ref{Exp:Simulink}. Local Area Network (LAN) connection is used for communication between the two PCs.
Crazyradio is a 2.4 GHz long range USB radio dongle, which is used to communicate with the crazyflie. 

\begin{figure}[htbp]
	\centering
	\includegraphics[width=9.25cm,trim={0.0cm 0.5cm 0.0cm 1.5cm},clip]{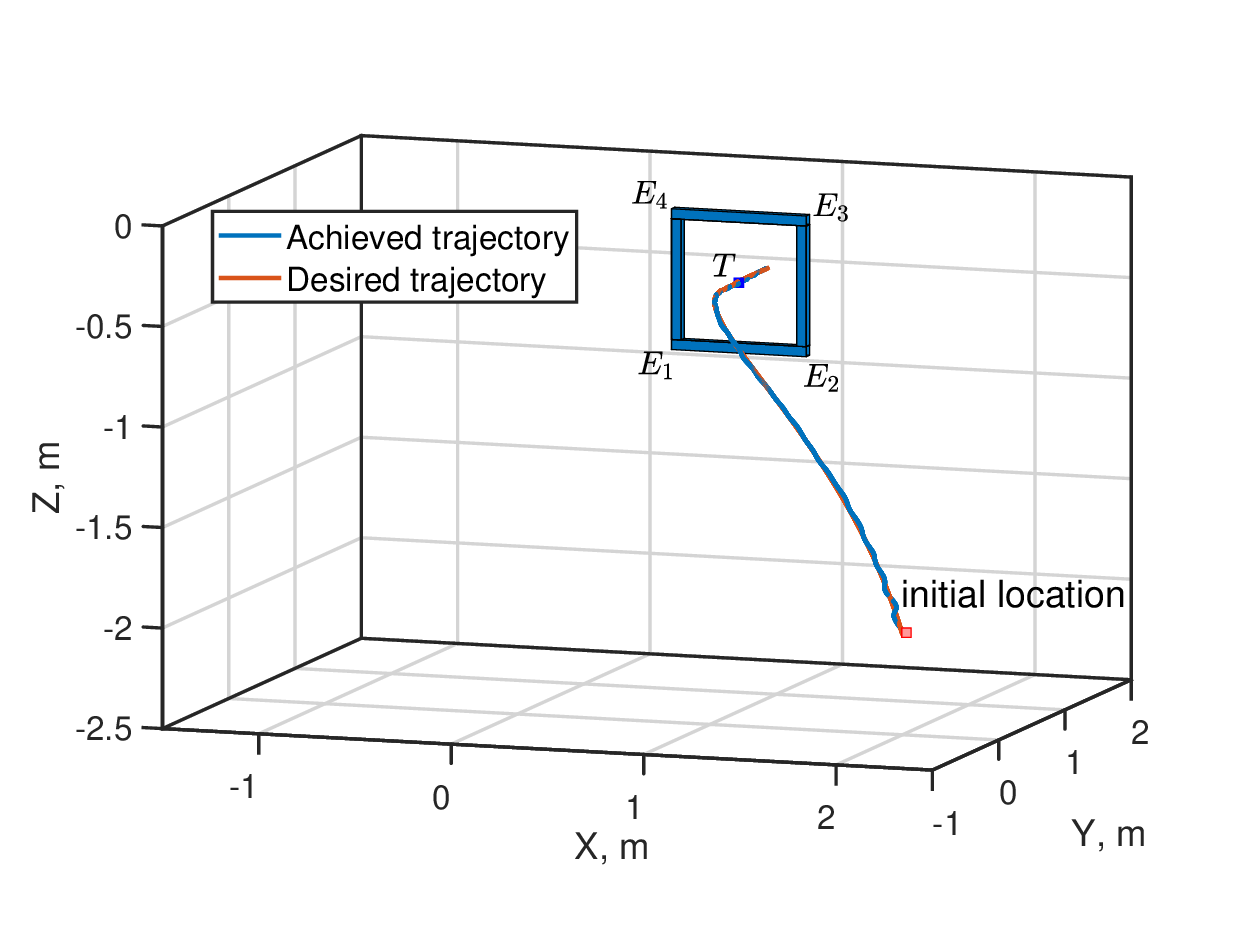}
	\caption{Quadrotor trajectory.}
	\label{Exp:windowtrajectory}
\end{figure}

The window presents a gap of dimension $0.60\times0.60$ m with its four vertices are located at $E_1(0.43, 1.22, -0.8)$ m, $E_2(1.03, 1.22, -0.8)$ m, $E_3(1.03,1.22,-0.2)$ m, and $E_4(0.43,1.22,-0.2)$ m, as shown in Fig. \ref{Exp:windowtrajectory}. 
\begin{figure}[htbp] 
	\centering
	\subfloat[{}]
	{\includegraphics[width=7.75cm]{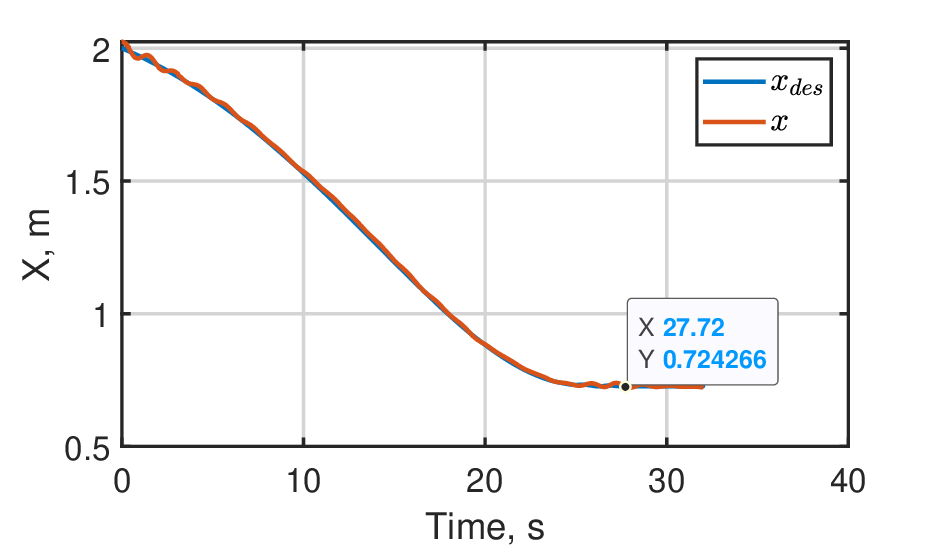}}\\
	\subfloat[{}]
	{\includegraphics[width=7.75cm]{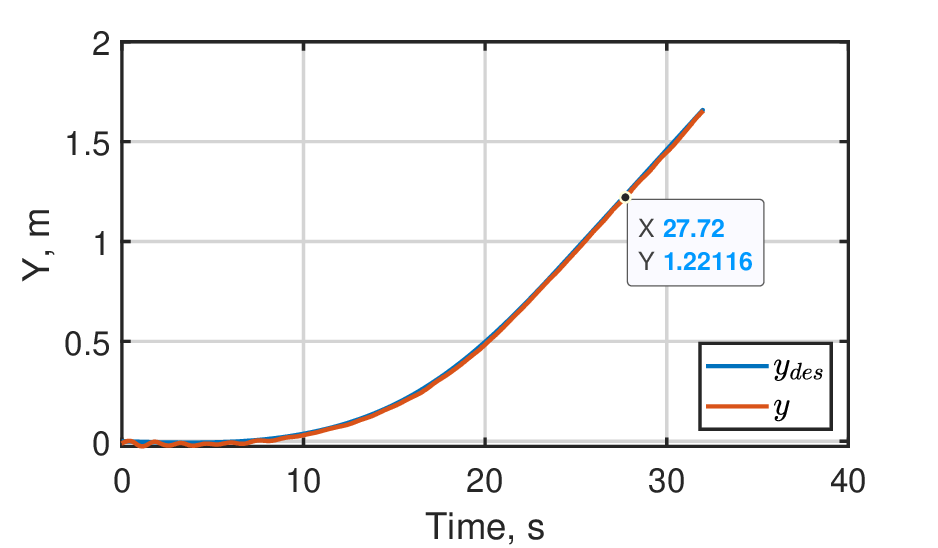}}\\
	\subfloat[{}]
	{\includegraphics[width=7.75cm]{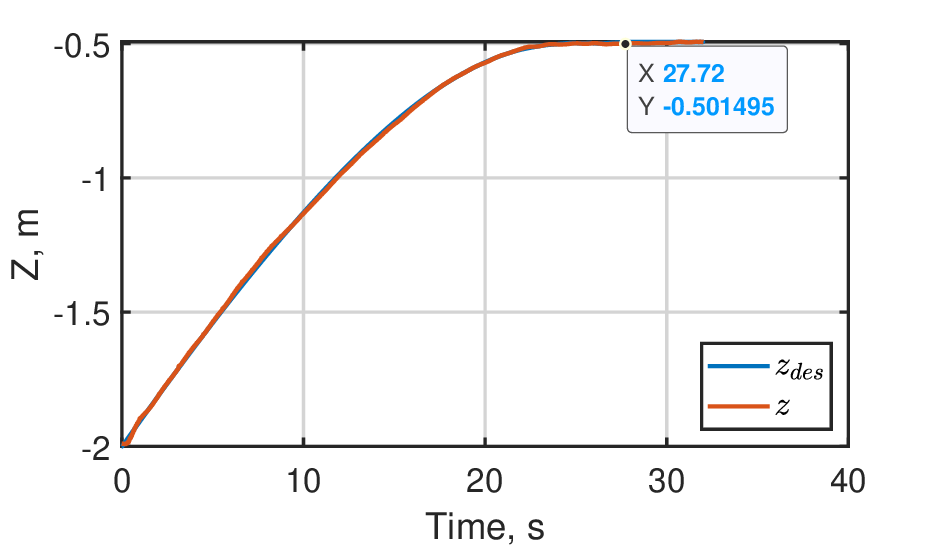}}
	\caption{Trajectory components. (a) X component. (b) Y component. (c) Z component.}
	\label{Trajectory components}
\end{figure}
The initial location for the Crazyflie is set to be at $(2,0,-2)$ m. The experiment is carried out with the commanded yaw rate and commanded yaw angle set to zero. The instantaneous azimuth angles $\beta_i$'s and elevation angles $\alpha_i$'s (for $i=1,2,3,4$) of the four vertices are evaluated in the ``Traversal Guidance block'' shown in Fig. \ref{Exp:Simulink} using the instantaneous position of the Crazyflie and window vertices obtained through motion capture system. This information is then utilized to generate the guidance commands $\gamma_{des}$ and $\chi_{des}$ as in Eqs. \eqref{desg} and \eqref{desheading}. Once the traversal condition \eqref{c2} is met, the guidance commands follow their already achieved heading angle $\chi_{des}=\frac{\pi}{2}$ rad and flight path angle $\gamma_{des}=0$ rad. Utilizing Eqs. \eqref{trajectory}-\eqref{zdes} the desired velocity components $(\dot{x}_{des}$, $\dot{y}_{des}$, $\dot{z}_{des})$ and the desired position components $({x_{des}}$, ${y_{des}}$, ${z_{des}})$ are generated in the ``Traversal Guidance Block'' shown in Fig. \ref{Exp:Simulink}. This experimental study considers the traversal velocity of Crazyflie as $V=0.1$ m/s.
The desired trajectory generated by the guidance algorithm and the achieved trajectory of the qudrotor are shown in Fig. \ref{Exp:windowtrajectory}. 
\begin{figure}[htbp] 
	\centering
	\subfloat[{}\label{Exp:Elevation}]
	{\includegraphics[width=7.75cm]{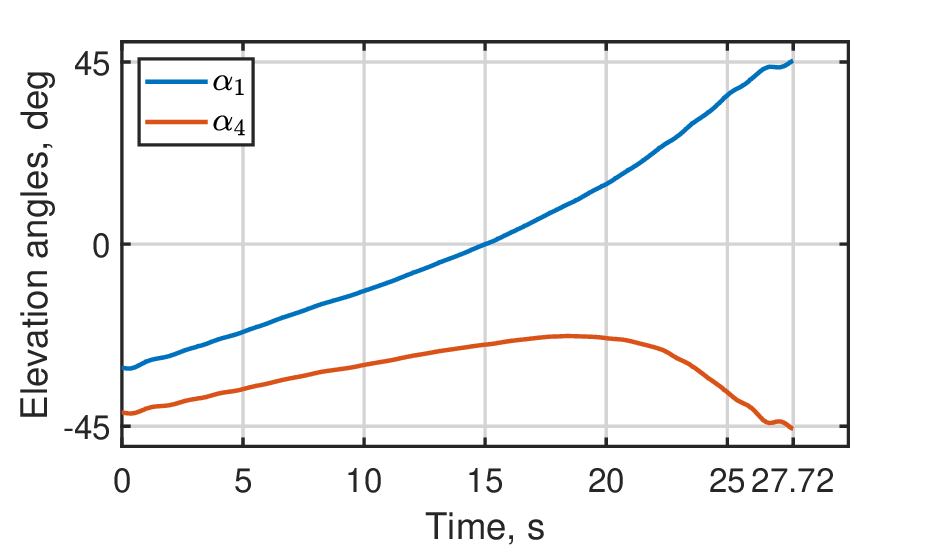}}\\
	\subfloat[{}\label{Exp: azimuth}]
	{\includegraphics[width=7.75cm]{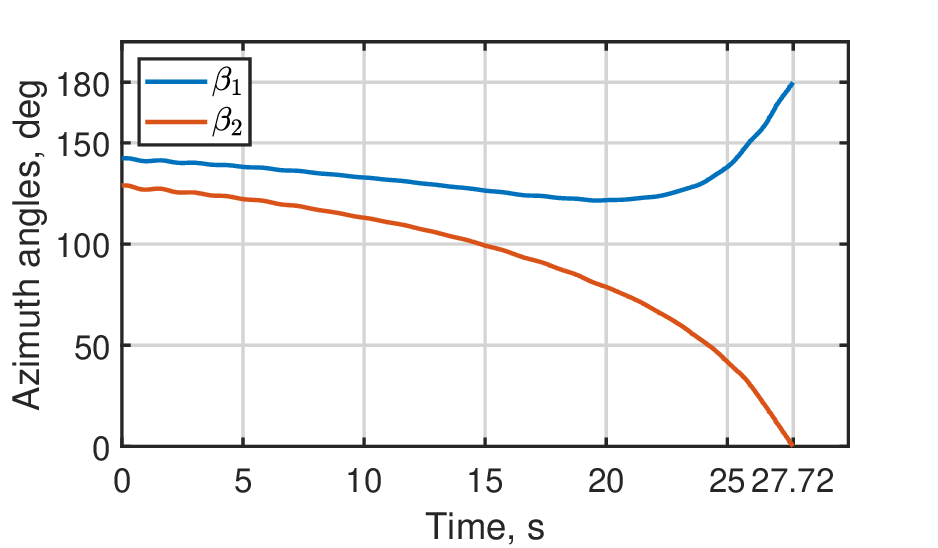}}
	\caption{Bearing angles for guidance computation: (a) Elevation angles. (b) Azimuth angles.}
\end{figure}	
\begin{figure}[htbp] 
	\centering
	\subfloat[{}\label{Exp:bisector1}]
	{\includegraphics[width=7.780cm]{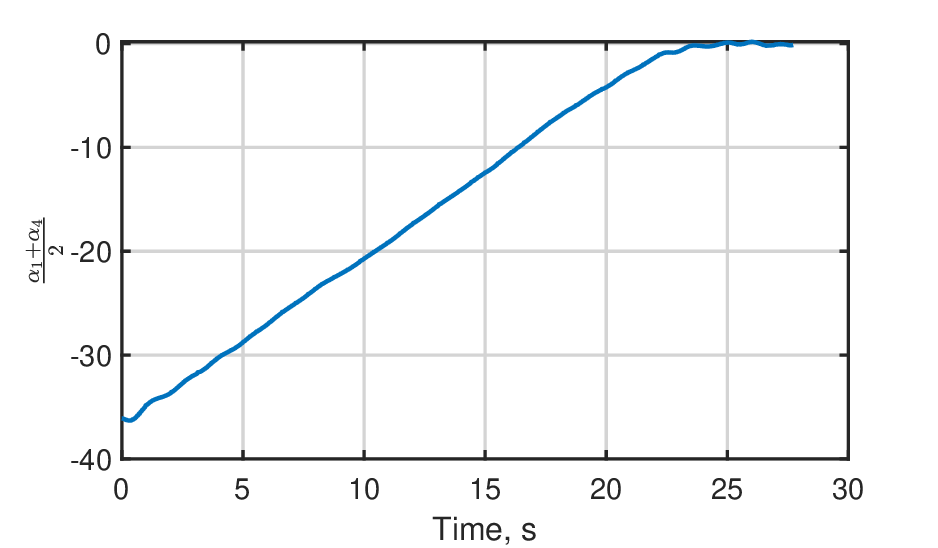}}\\
	\subfloat[{}\label{Exp:bisector2}]
	{\includegraphics[width=7.780cm]{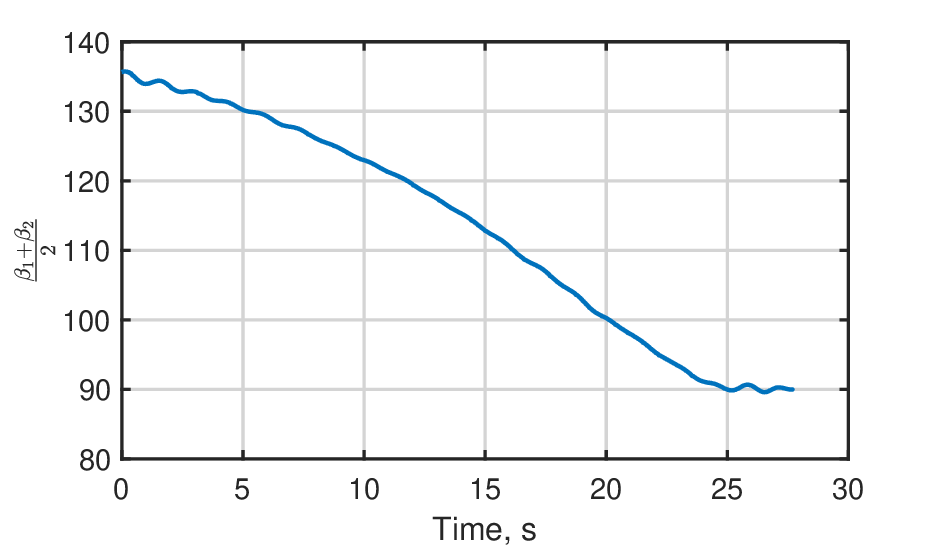}}\\
	\subfloat[{}\label{Exp:S1}]
	{\includegraphics[width=7.780cm]{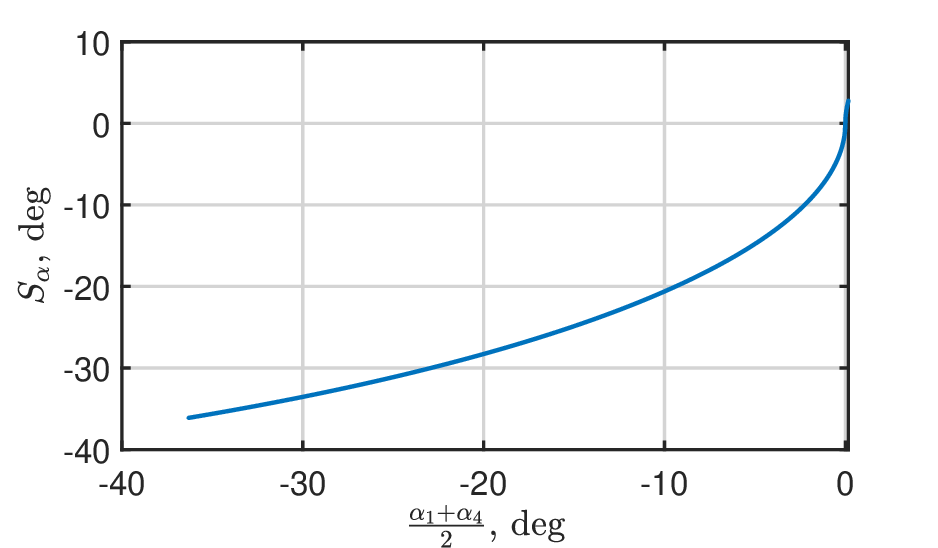}}\\
	\subfloat[{}\label{Exp:S2}]
	{\includegraphics[width=7.780cm]{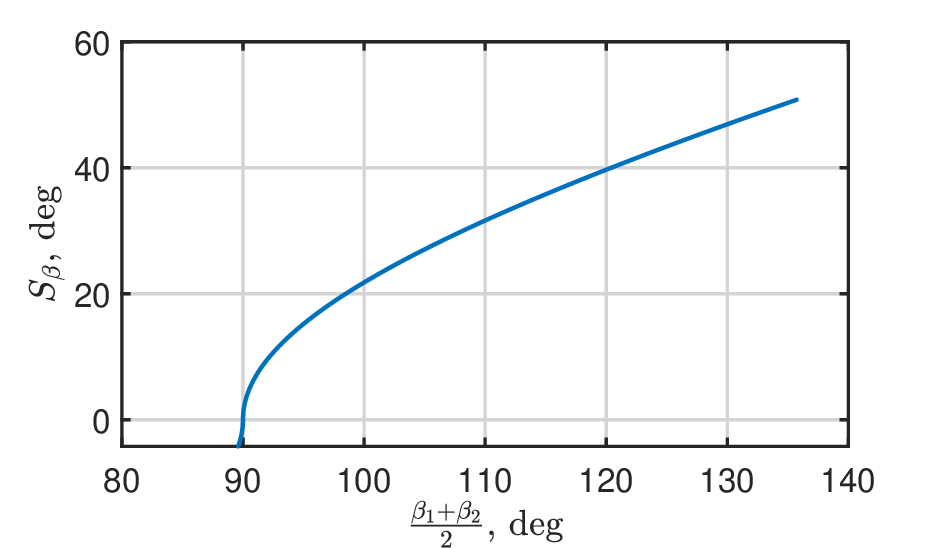}}\\
	\caption{Angular components for guidance computation: (a) Bisector of elevation angles. (b) Bisector of azimuth angles. (c) Shaping angle component for proposed flight path angle. (d) Shaping angle component for proposed heading angle. }
\end{figure}
It is observed that the quadrotor tracks the commanded trajectory with a maximum error of less than $0.0283$ m. From Fig. \ref{Trajectory components}, it can be seen that the quadrotor successfully reaches the traversal point $T(0.7242,1.2216,-0.5014)$ m at $27.72$ s, with a negligible error of 0.014 m from the centroid of the window. The elevation angles with respect to the vertices $E_1$ and $E_4$, and the azimuth angles with respect to $E_1$ and $E_2$ are depicted in Figs. \ref{Exp:Elevation} and \ref{Exp: azimuth}, respectively. The elevation angles and azimuth angles achieve their desired values as $(\alpha_1,\alpha_3)=(45,-45)$ deg and $(\beta_1,\beta_2)=(180,0)$ deg, respectively. 
From Figs. \ref{Exp:bisector1} and \ref{Exp:bisector2}, it can be seen that the bisector of elevation angles $\frac{\alpha_1+\alpha_3}{2}$ and the bisector of azimuth angles $\frac{\beta_1+\beta_2}{2}$ attain their desired value of $0$ deg and $90$ deg, respectively, as the quadrotor approaches the centroid of the window. The shaping angle requirements monotonically diminish to zero, as depicted in Figs. \ref{Exp:S1}  and \ref{Exp:S2}. 	
\section{CONCLUSION}\label{Conclusion}
 This work presented a novel guidance solution for three dimensional rectangular window traversal problem by a quadrotor. By utilizing the bearings-only information of window vertices, the proposed guidance logic governs the commanded flight path and heading angles of the vehicle, incorporating an angular bisector component and an elliptical shaping component, which enables lateral traversal through the window. 
 The quadrotor motion is shown to asymptotically converges to the line, which is normal to the window plane, and passes through its centroid. The proposed guidance commands are simple, expressed in closed form, and easy to compute. A comparison with the state-of-the-art methods highlights that, in contrast to either complete prior information of the window or its depth information, the proposed method requires only the relative bearing angle information of the window vertices. Simulation results highlight the effectiveness of the proposed guidance solution considering several initial conditions, dynamic limits on quadrotor attitude angles, and noisy bearing measurements. Additionally, Monte Carlo studies, which incorporate normally distributed noise in the bearing information,
emphasize the robustness of the proposed method. Experimental flight validation trials are carried out using an indoor motion capture system and a Crazyflie 2.0 quadrotor.

The proposed guidance commands are angular inputs, and accordingly, they do not inherently impose any speed or acceleration requirements on the vehicle. Moreover, the traversal trajectories are dynamically generated based on instantaneous bearing angles, enabling the flexibility of incorporating updated information as the flight progresses. 
Considering the camera field-of-view limits or the presence of obstacles can be motivation for potential future works. \vspace*{-.5pc}

\bibliographystyle{IEEEtran}
\bibliography{newsample}

\end{document}